%% file: main.tex
\documentclass{article}
\usepackage{ijcai26}

\usepackage{times}
\usepackage{soul}
\usepackage{url}
\usepackage[hidelinks]{hyperref}
\usepackage[utf8]{inputenc}
\usepackage[small]{caption}
\usepackage{graphicx}
\usepackage{amsmath}
\usepackage{amsthm}
\usepackage{booktabs}
\usepackage{algorithm}
\usepackage{algorithmic}
\usepackage[switch]{lineno}

\newtheorem{example}{Example}

\newtheorem{definition}{Definition}
\newtheorem{assumptions}{Assumption}
\theoremstyle{remark}

\usepackage[babel=true,expansion=true,kerning=true,tracking=true,spacing=true,verbose=silent]{microtype} 
\usepackage{inconsolata} 
\usepackage{csquotes} 
\usepackage{graphbox} 
\usepackage{multirow}

\usepackage[capitalize,nameinlink,noabbrev]{cleveref}
\crefname{section}{Sec.}{Secs.}
\crefname{assumptions}{Ass.}{Ass.}
\usepackage{enumitem}
\newlist{questions}{enumerate}{1}
\setlist[questions]{label=Q\arabic*:, ref=Q\arabic*, leftmargin=*}
\crefformat{questionsi}{#2RQ#1#3}
\Crefformat{questionsi}{#2RQ#1#3}
\crefmultiformat{questionsi}{#2RQ#1#3}{ and #2RQ#1#3}{, #2RQ#1#3}{, and #2RQ#1#3}
\Crefmultiformat{questionsi}{#2RQ#1#3}{ and #2RQ#1#3}{, #2RQ#1#3}{, and #2RQ#1#3}
\crefrangeformat{questionsi}{#3RQ#1#4--#5RQ#2#6}
\Crefrangeformat{questionsi}{#3RQ#1#4--#5RQ#2#6}
\usepackage{mathtools,amssymb,dsfont}
\usepackage{relsize}
\usepackage{siunitx}
\usepackage{subcaption}
\usepackage{makecell}
\usepackage[dvipsnames]{xcolor}
\usepackage{xspace}

\usepackage[toc,page,header]{appendix}
\usepackage{minitoc}

\newcommand*{\compressedparagraph}[1]{\par\noindent\textbf{#1}}

\newcommand{\Concept}[1]{\ensuremath{\text{\normalfont\texttt{#1}}}}
\newcommand{\Ontology}{\ensuremath{\mathcal{O}}\xspace}

\DeclareMathOperator*{\mean}{mean}

\DeclareMathOperator*{\argmin}{arg\,min}

\usepackage{forest} 
\colorlet{leaf}{blue!80!black}
\tikzset{
  my node/.style={
    draw=gray,
    thin,
    rounded corners=3,
    text height=1.3ex,
    text depth=0ex,
    font=\sffamily,
    inner sep=2pt,
    text centered,
  },
  root node/.style={
    my node,
    draw=none,
    font=\itshape{root},
  },
  leaf node/.style={
    my node,
    draw=leaf,
    font=\sffamily\color{leaf},
  },
  strange node/.style={
    my node,
    inner color=red!5, outer color=red!80,
  },
}
\forestset{
  default preamble={
    for tree={
      my node,
      l=13pt,
      l sep=0, s sep=5pt,
    },
  }
}

\title{%
  Explaining, Verifying, and Aligning Semantic Hierarchies in Vision-Language Model Embeddings
}

\author{
  Gesina Schwalbe$^{1,2}$\and
  Mert Keser$^{3,4}$\and
  Moritz Bayerkuhnlein$^2$\and
  Edgar Heinert$^{5}$\and
  Annika Mütze$^{5}$\and
  Marvin Keller$^2$\and
  Sparsh Tiwari$^2$\and
  Georgii Mikriukov$^{6}$\and
  Diedrich Wolter$^2$\and
  Jae Hee Lee$^{7}$\And
  Matthias Rottmann$^{5}$
  \\
  \affiliations
  $^1$Ulm University, Germany\\
  $^2$University of Lübeck, Germany\\
  $^3$Technical University of Munich, Germany\\
  $^4$AUMOVIO SE, Germany\\
  $^5$Osnabrück University, Germany\\
  $^6$Leibniz Institute for Agricultural Engineering and Bioeconomy, Germany\\
  $^7$University of Hamburg, Germany\\
  \emails
  gesina.schwalbe@uni-ulm.de,
  \{moritz.bayerkuhnlein, sparsh.tiwari, diedrich.wolter\}@uni-luebeck.de,
  mert.keser@aumovio.com,
  \{edgar.heinert, annika.muetze, matthias.rottmann\}@uni-osnabrueck.de,
  marvin.keller@student.uni-luebeck.de,
  gmikriukov@atb-potsdam.de,
  jae.hee.lee@uni-hamburg.de
}

\begin{document}

\maketitle

\begin{NoHyper}
  \stepcounter{footnote}
  \footnotetext{Find supplementary material at: \url{https://arxiv.org/abs/2603.26798}}
  \stepcounter{footnote}
  \footnotetext{Find code at: \url{https://github.com/gesina/ontological-commitment}}
\end{NoHyper}

\begin{abstract} 
  Vision-language model (VLM) encoders such as CLIP enable strong retrieval and zero-shot classification in a shared image--text embedding space, yet the semantic organization of this space is rarely inspected. We present a post-hoc framework to \emph{explain}, \emph{verify}, and \emph{align} the semantic hierarchies induced by a VLM over a given set of child classes. First, we extract a binary hierarchy by agglomerative clustering of class centroids and name internal nodes by dictionary-based matching to a concept bank. Second, we quantify plausibility by comparing the extracted tree against human ontologies using efficient tree- and edge-level consistency measures, and we evaluate utility via explainable hierarchical tree-traversal inference with uncertainty-aware early stopping (UAES). Third, we propose an ontology-guided post-hoc alignment method that learns a lightweight embedding-space transformation, using UMAP to generate target neighborhoods from a desired hierarchy. Across 13 pretrained VLMs and 4 image datasets, our method finds systematic modality differences: image encoders are more discriminative, while text encoders induce hierarchies that better match human taxonomies. Overall, the results reveal a persistent trade-off between zero-shot accuracy and ontological plausibility and suggest practical routes to improve semantic alignment in shared embedding spaces.
\end{abstract}

\section{Introduction}

\newcommand{\myone}{\includegraphics[width=2ex]{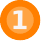}}
\newcommand{\mytwo}{\includegraphics[width=2ex]{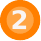}}
\newcommand{\mythree}{\includegraphics[width=2ex]{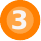}}
\newcommand{\myfour}{\includegraphics[width=2ex]{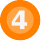}}
\newcommand{\myfive}{\includegraphics[width=2ex]{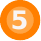}}
\begin{figure}[t]
  \centering
  \hspace*{-0.07\linewidth}\includegraphics[width=1.1\linewidth]{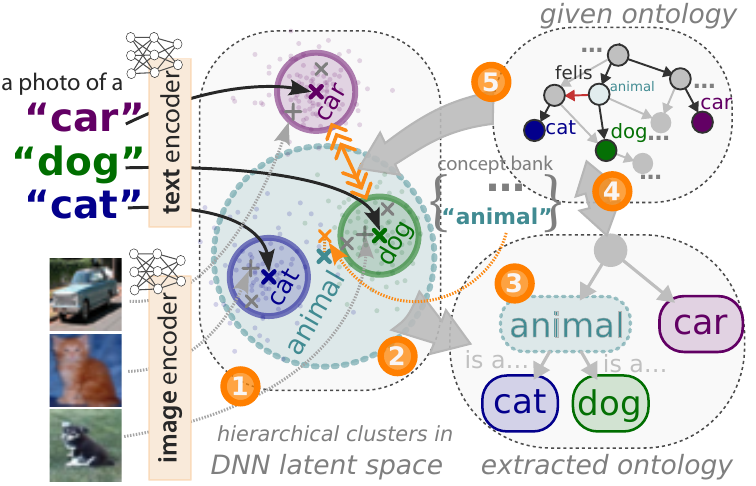}
  \caption{Our post-hoc framework to \textbf{explain} (embed \myone{} given concepts, cluster~\mytwo{} and name found parents \mythree{}), \textbf{verify} \myfour{}, and \textbf{align} \myfive{} the semantic hierarchy induced by a VLM in its embedding space.}
  \label{fig:overview}
\end{figure}

Contrastive vision-language models (VLMs) such as CLIP Radford et al.~\shortcite{radford2021learning} demonstrate remarkable capabilities in learning joint representations of images and text \cite{%
  barraco
}.
These representations are adopted in various applications, including image retrieval~\cite{baldrati2022effective}, captioning~\cite{%
  barraco%
},
and zero-shot classification~\cite{radford2021learning}.
However, their evaluation is still dominated by task-level performance metrics such as classification accuracy that provide little insight into how concepts are internally organized and related.
In particular, understanding the semantic hierarchies of these representations has not been thoroughly explored \cite{lee2025conceptbased}.

Semantic hierarchies capture how concepts relate via commonalities and differences, and are central to human categorization and symbolic knowledge representation via ontologies~\cite{confalonieri_multiple_2025,guarino1998formal}.
Extracting the hierarchy induced by a VLM is crucial for explainability~\cite{bhalla_interpreting_2024,alper_emergent_2025,lee2025conceptbased,%
  wan2020nbdt
}. It further enables identifying unintuitive or biased groupings (e.g., grouping \Concept{man} and \Concept{woman} primarily via \Concept{hair} rather than under \Concept{person}) and aligning learned similarities with human knowledge~\cite{ribeiro2021aligning,confalonieri_multiple_2025}.
Lastly, hierarchies provide an inductive bias for more robust and informative classification~\cite{%
  bertinetto_making_2020,%
  dhall_hierarchical_2020,%
  ge2023improving%
}, such as explainable hierarchical tree-traversal \cite{wan2020nbdt}.

Meanwhile, key desiderata formulated for VLM embedding spaces are text-to-image alignment, and that classes can be \emph{discriminated} via zero-shot classification.
We complement this by two new desiderata: Text and image encoders should also align in their induced hierarchies; and they should be \emph{plausible}, i.e., the hierarchy should match human taxonomies.
Note that plausibility and discriminability may diverge, since embedding similarities may accurately discriminate class members, but still violate human expectations.

Into that direction, prior work has explored the interpretability of VLM embeddings~\cite{%
  bhalla_interpreting_2024
} (\emph{Which concepts are encoded?}) and found and reinforced pre-defined semantic hierarchies in them ~\cite{alper_emergent_2025} (\emph{Is a given hierarchy encoded?}).
However, there is a lack of systematic methods to answer the following questions for a trained VLM encoder, along which we organize our pipeline (cf.\ \cref{fig:overview}):
\begin{questions}[wide,listparindent=0pt, labelindent=0pt]
\item\label{rq1} Can we \textbf{explain} \emph{which semantic hierarchy} is induced by an encoder for a given set of leaf classes? How faithful and reliable are local explanations via explainable hierarchical inference? (\cref{sec:extraction,sec:tree-traversal,sec:faithfulness})
\item\label{rq2} Can we \textbf{verify} the plausibility of induced hierarchies relative to reference ontologies? How do they vary across modalities, models, and hierarchy depths? (\cref{sec:verify,sec:plausibility})
\item\label{rq3} Can we post-hoc \textbf{align} an induced hierarchy to a target similarity structure without sacrificing zero-shot accuracy? (\cref{sec:align,sec:exp-alignment})
\end{questions}

\paragraph{Contributions.}
We demonstrate the utility of semantic hierarchy extraction:
\begin{itemize}[label=$\blacktriangleright$, leftmargin=1em]
  \item We propose a post-hoc pipeline to explain, verify, and align semantic hierarchies in VLM embedding spaces.
    We extract a hierarchy via hierarchical clustering of leaf classes and dictionary-based parent naming, verify it with efficient ontology-based comparison metrics, and align it using UMAP-guided target neighborhoods and a lightweight embedding transformation.
  \item We empirically study 13 pretrained VLMs across 4 datasets and multiple common-sense ontologies, enabling systematic comparison of encoder semantics.
  \item We find a consistent modality gap: image encoders yield higher zero-shot accuracy, while text encoders induce more plausible hierarchies; and enriching hierarchical inference with UAES improves semantic correctness, bridging the explainability-faithfulness trade-off.
\end{itemize}

\begin{figure*}[t]
  \begin{subfigure}{.75\linewidth}
    \resizebox{.98\linewidth}{!}{%
      \centering\scriptsize
      \begin{forest}
        [ siglip: image, root node
          [ fleet
            [ starboard
              [ airplane, leaf node ]
            [ ship, leaf node ]]
            [ vehicular
              [ car, leaf node ]
          [ truck, leaf node ]]]
          [ warm-blooded
            [ horse, leaf node ]
            [ doglike
              [ dog-sized
                [ cat, leaf node ]
              [ dog, leaf node ]]
              [ croaky
                [ frog, leaf node ]
                [ avifaunal
                  [ bird, leaf node ]
        [ deer, leaf node ]]]]]]
      \end{forest}
      \hspace*{-2em}\raisebox{1.75\baselineskip}{%
        \begin{forest}
          [ siglip: text and image, root node
            [ animal
              [ avifaunal
                [ bird, leaf node ]
              [ frog, leaf node ]]
              [ canine
                [ domestic cat
                  [ cat, leaf node ]
                [ dog, leaf node ]]
                [ equine
                  [ deer, leaf node ]
            [ horse, leaf node ]]]]
            [ vehicular
              [ motor vehicle
                [ car, leaf node ]
              [ truck, leaf node ]]
              [ boat
                [ airplane, leaf node ]
          [ ship, leaf node ]]]]
        \end{forest}
      }
    }%
    \caption{Comparison of two extracted hierarchies over the CIFAR-10 leaves (shown in \textcolor{leaf}{blue}) from the SigLIP model, using image-only leaf embeddings (left) versus combined text\&image leaf embeddings (right).}\label{fig:example-commitment}%
  \end{subfigure}%
  \begin{subfigure}{.25\linewidth}
    \centering
    \includegraphics[width=\linewidth, align=b]{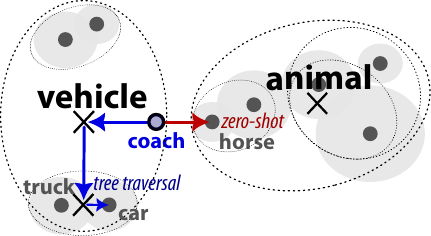}%
    \caption{Illustration of tree-traversal vs.\ zero-shot inference in embedding space of the text \& image tree.}\label{fig:tree-traversal}
  \end{subfigure}
  \caption{
    Running toy example on CIFAR-10 leaf classes.
  }
\end{figure*}

\section{Related Work}

\paragraph{Explaining Learned Concepts.}
The field of concept-based XAI (c-XAI) has evolved around the goal to associate vectors in the latent representations with semantic concepts \cite{%
  lee2025conceptbased
}.
Such associations have found manifold applications in the inspection of a DNN's learned ontology, such as:
which concepts from a given ontology are learned? \cite{%
  ribeiro2021aligning%
}, how similar are pairs of concepts? \cite{%
  fong2018net2vec,
  schwalbe2021verification
}, and how are concepts from different layers or models correlated?
\cite{%
  kim2018interpretability,
  mikriukov2023revealing
}.
Notably, we here build upon the approach by Yuksekgonul et al.~\shortcite{yuksekgonul2022posthoc} to name parent embeddings.
They utilize the text-to-image alignment in VLMs to match vectors to the closest text embeddings from a dictionary.
To go from concepts to hierarchical concept relations, Wan et al.~\shortcite{wan2020nbdt} employed hierarchical clustering on latent representations, but unlike our approach restricted to the last layer of a classifier.
And despite rising interest in verification and steering of DNNs against ontologies
\cite{%
  ribeiro2021aligning,
  ponomarev2022ontology,
  badredd,
  alper_emergent_2025%
},
it remains largely unexplored how to post-hoc explain and control the learned semantic similarity structure within DNN embedding spaces \cite{%
  confalonieri_multiple_2025
}.

\paragraph{Understanding VLM Embedding Spaces.}
The CLIP-based learning scheme \cite{%
  radford2021learning,
  jia2021Scalingvisual
} of concurrent VLMs have been shown to emerge deep concept understanding \cite{%
  bhalla_interpreting_2024,
  zhao_quantifying_2025
}, and even class hierarchies \cite{alper_emergent_2025,ge2023improving}—though existing hierarchy extraction techniques rely on costly parent supervision and don't align with ontology concept dictionaries.
Moreover, previous findings that integrating hierarchy information into the inference process makes classification more explainable \cite{wan2020nbdt} and reliable \cite{%
  bertinetto_making_2020,%
  dhall_hierarchical_2020%
}, were shown to transfer to VLMs \cite{%
  novack_chils_2023
}.
We therefore adapt the hierarchy tree-traversal approach from Wan et al.~\shortcite{wan2020nbdt} for VLM classification, newly adding early stopping.
Despite these impressive properties, the modality alignment in CLIP embedding spaces remains suboptimal, mapping text and image to distinct latent regions \cite{liang_mind_2022,eslami_mitigate_2024}, and requiring prompt finetuning to boost zero-shot performance \cite{radford2021learning,ma2025does,ge2023improving}.
Other than prior studies, we here for the first time investigate the text-to-image alignment with respect to the primarily learned semantic hierarchies of the encoders.

\paragraph{Ontologies and Ontology Matching.} 
An ontology reflects a perspective on a domain \cite{guarino1998formal}, formalized as concepts, relations, and assumptions thereon \cite{guarino1998formal}.
By now a variety of both specialized and general domain ontologies are available \cite{confalonieri_multiple_2025,mascardi2007comparison}. The latter include closed-source ones \cite{lenat1989building,forbus2017analogy}, large-scale but less principled ones like ConceptNet \cite{speer2017conceptnet} or DBPedia \cite{lehmann}.
We here use the well-established, large, general domain ontologies SUMO~\cite{niles2001standard}, OpenCyc~\cite{Matuszek}, and Yago~\cite{YAGO3AK} as reference for human knowledge on plausible hierarchies.
Ontology matching techniques aim to compare or integrate divergent ontologies \cite{OTEROCERDEIRA2015949}. A common distance measure for graphs is \emph{edit distance}, which transforms one instance into another via minimal edit operations \cite{%
  zhang1989simple,
  kutsu2010tree
}.
For our special case of semantic hierarchy trees, tree edit distance measures can be applied \cite{zhang1992editing}.
Aside from this global comparison, we here also aggregate local edge-wise scores based on matching lowest common ancestors (LCAs) \cite{bender2005lowest}.
However, existing works have not yet used ontology matching techniques to assess DNN-learned ontologies.

\section{Background}\label{sec:background}
The goal of our approach is to discover a learned semantic hierarchy, constituted of latent embeddings of parent concepts.
In the following we formalize semantic hierarchies, and recap the classical techniques for concept discovery as well as underlying key assumptions for hierarchy extraction.

\subsection{Ontologies}
A semantic hierarchy is a slice of a broader ontology.
\begin{definition}[Ontology]\label{def:ontology}
  An \emph{ontology} \Ontology is a description of a domain by a set of class membership functions (the \emph{concepts}), instance \emph{relations}, and a logic theory of assumptions thereon.
  \cite{guarino1998formal}
\end{definition}
Semantic similarity is an example for relations between instances.
This paper only considers taxonomic \emph{is-a} assumptions on concepts (hypernymy/subclass relations \cite{fellbaum2010wordnet}) for verification and alignment. We denote these as $\Concept{IsParentOf}(\Concept{P},\Concept{C})\coloneqq (\forall v\colon \Concept{C}(v) \Rightarrow \Concept{P}(v))$), meaning that instances of \Concept{C} are also instances of \Concept{P} (e.g., \Concept{cat} $\Rightarrow$ \Concept{animal}).
A semantic hierarchy is a tree formed by such relations; our goal is to extract such a hierarchy from a VLM's embedding space.

\subsection{Assumed Properties of VLM Latent Spaces}
We rely on a few high-level assumptions that are standard in concept-based interpretability \cite{lee2025conceptbased},
where often single vectors are used as prototypes for concepts \cite{kim2018interpretability,fong2018net2vec} (\cref{ass:embeddability}); and
in the typical VLM training and zero-shot inference schemes which employ cosine similarity as a proxy for semantic similarity of latent representations (\cref{ass:interpolatability}). These assumptions are formalized in the following.

First, we simplify the representation of a visual concept from its cluster of concept instance embeddings to a point in embedding space.
\begin{assumptions}[Concept Embeddability]\label{ass:embeddability}
A human-interpretable visual concept $\Concept{C}$ can be represented by a point $e(\Concept{C})$, its \emph{concept embedding}, in the VLM embedding space \cite{kim2018interpretability,fong2018net2vec}.
\end{assumptions}
Notable work introducing and validating this modeling scheme of concepts as single vectors were the concept activation vectors by Kim et al.~\shortcite{kim2018interpretability} and concept embeddings by Fong et al.~\shortcite{fong2018net2vec}. More recently, Mikriukov et al.~\shortcite{mikriukov2025local} confirmed that despite the more distributed nature of concept instance representations, single vectors are usually a good approximation.

Second, cosine similarity between concept embeddings is a reasonable proxy for semantic relatedness, which is our key to find parent embeddings. This also is the same assumption used by zero-shot classification where we assign $x$ to the closest concept embedding under cosine distance $d(u,v) \coloneqq 1 - \frac{u\cdot v^T}{\|u\|\cdot\|v\|}$ in embedding space \cite{radford2021learning}.
\begin{assumptions}[Embedding Interpolatability]\label{ass:interpolatability}
A concept $\Concept{C}$ is more similar to $\Concept{C}'$ than to $\Concept{C}''$
iff $d(e(\Concept{C}),e(\Concept{C}'))\leq d(e(\Concept{C}),e(\Concept{C}''))$
\cite{%
  fong2018net2vec,%
  ghorbani2019automatic,
  posada-moreno2024eclad
}.
\end{assumptions}
This interpolatability property was empirically confirmed for concept embeddings both in language \cite{mikolov2013linguistic} and vision model latent spaces \cite{fong2018net2vec,ghorbani2019automatic,posada-moreno2024eclad}.

If cosine similarity is used to assign an instance's embedding $v$ to a concept $\Concept{C}$ ($\Concept{C}(v)$ from above), a parent concept for given children concepts is the one most similar to all instances of its children simultaneously.
Hence, a direct consequence of \cref{ass:interpolatability} is to assume that DNNs learn hierarchies in terms of similarity structures: Concept vectors of parent concepts can be approximated by aggregating their children's concept vectors (e.g., by averaging \cite{%
  mikriukov2025local,
  wan2020nbdt
}), which motivates our use of agglomerative clustering.

\section{Methodology}
This section introduces our pipeline to \emph{explain} the semantic hierarchy induced by a VLM as binary tree of concepts (\cref{sec:extraction}) and via enhanced explainable hierarchical classification (\cref{sec:tree-traversal}),
which allows to \textbf{verify} the VLM plausibility against references
(\cref{sec:verification}), and semantically \textbf{align} its embedding space to a target similarity structure (\cref{sec:align}).

\begin{example}\label{ex:running-cifar10}
To make the pipeline concrete, we use the CIFAR-10 leaf classes (airplane, automobile, bird, cat, deer, dog, frog, horse, ship, truck) as a running example throughout the paper.
\Cref{fig:example-commitment} shows two hierarchies extracted for the same ten leaves from SigLIP, and foreshadows the modality-dependent differences we quantify later (\cref{sec:modality}).
\end{example}

\compressedparagraph{Experimental Setup and Notation.}
We assume a trained VLM encoder $F$ and a dataset of leaf classes (e.g., CIFAR-10 \cite{alex2009learning}) with train/test splits. Leaf concepts can be represented via images, text descriptions (e.g., LLM-generated prompts \cite{ma2025does}), or both. We further assume a concept bank $\mathcal{C}$ of candidate parent concepts, also with text descriptions, that map to a given reference ontology \Ontology. For $\mathcal{C}$  we here consider the large-scale lexical database WordNet \cite{fellbaum2010wordnet}, which conveniently maps to many general domain ontologies like SUMO \cite{niles2001standard}.

\subsection{Hierarchy Extraction}
\label{sec:extraction}
We aim to extract a labeled binary tree $T$ that summarizes in a human-inspectable manner how the encoder groups the given leaf classes in embedding space.
Leaves correspond to the leaf dataset classes; each internal node represents an extracted parent concept from the concept bank together with an embedding.
Edges represent predicted $\Concept{IsParentOf}$ relations (cf.\ \cref{fig:example-commitment}).

\begin{example}[Parent Relations]\label{ex:running-cifar10-explain}
In \cref{fig:example-commitment}, image-only leaf embeddings group \Concept{bird} and \Concept{deer} under \Concept{avifaunal} as captured by the extracted binary tree structure.
\end{example}

\paragraph{Approach.}
We build $T$ in three steps, as shown in \cref{alg:hierarchy-extraction}.
First, we compute a centroid embedding $e(\Concept{C})$ for each leaf class \Concept{C} from embeddings of train samples for $\Concept{C}$ (image, text, or both).
Second, we run \emph{agglomerative hierarchical clustering} \cite{wardjr.1963hierarchical} on the leaf centroids using cosine similarity and average linkage; each merge creates a parent node whose embedding is the normalized mean of its children's embeddings.
Third, we name internal nodes by matching each parent embedding $p$ to the closest concept-bank embedding $e(\tilde{\Concept{P}})$, using a one-to-one matching (linear sum assignment \cite{crouse2k16}, LSA) to optimally avoid duplicate names.

\paragraph{Rationale.}
Intuitively, each concept \Concept{C} is defined by the set of samples $x$ to which \Concept{C} applies. Via $F$ this translates to a \emph{cluster of embeddings} $F(x)$; for leaf concepts we approximate this cluster using the embeddings of their training samples.
By definition, a parent concept $\Concept{P}$ should cover the cluster of each of its children concepts $\Concept{C}$: If for all $x$ holds $\Concept{C}(x)\implies \Concept{P}(x)$, then $\{x\mid \Concept{C}(x)\}\subset \{x\mid\Concept{P}(x)\}$.
This motivates iteratively merging the most similar clusters, as achieved by the agglomerative hierarchical clustering. Choosing cosine similarity reflects the objective used in CLIP training, differing from Euclidean-linkage choices in prior work on clustering concept embeddings in neural network latent spaces \cite{wan2020nbdt,mikriukov2025local}.
Intuitively, a good parent embedding for a parent cluster should minimize the distances to all of its children equally; for cosine distance on normalized vectors, the mean of children centroids fulfills this, which we assume to be the children's concept embeddings (Ass.\,\ref{ass:interpolatability}):
\begin{gather}
e(\Concept{P})\approx \operatorname*{mean}_{C \in \text{Children}(P)} e(\Concept{C})
\label{eq:superclass-is-mean}
\end{gather}
Averaging additionally robustifies the assignment of concept representatives and robustness increases with the number of samples in the leaf class datasets. Since clustering is applied to those robust means, it naturally becomes robust against fluctuations in the leaf datasets with more samples. Empirically we at most observed swaps between nearest and second-nearest siblings, as confirmed in qualitative UMAP visualizations (refer to supplementary for details).

\begin{algorithm}[tb]
\caption{Hierarchy extraction and parent naming}\label{alg:hierarchy-extraction}
\begin{algorithmic}
  \REQUIRE encoder $F$, leaf concepts $\{\Concept{C}_i\}_{i=1}^k$ with specification sets $\text{Spec}(\Concept{C}_i)$ (image, text, or both), concept-bank descriptions $\text{Descs}$, cosine distance $d$
  \\[.6\baselineskip]
  \COMMENT{\emph{1. Obtain normalized leaf concept centroids}}
  \FOR{$i=1,\dots,k$}
  \STATE $\bar e_i \gets \mean_{x\in\text{Spec}(\Concept{C}_i)} F(x)$
  \STATE $e(\Concept{C}_i)\gets \bar e_i / \|\bar e_i\|$
  \ENDFOR
  \\[.6\baselineskip]
  \COMMENT{\emph{2. Recover hierarchy using cosine average linkage}}
  \STATE $T \gets \text{AgglomerativeClustering}($\\%
    \hspace{\algorithmicindent}\hspace*{2em}%
    $\{e(\Concept{C}_i)\}_{i=1}^k,\
    \text{metric}=d,\
  \text{linkage}=\text{average})$
  \FOR{each internal node $\Concept{P}$ of $T$ in bottom-up order}
  \STATE $\bar e_{\Concept{P}} \gets \mean_{\Concept{C}\in\text{Children}(\Concept{P})} e(\Concept{C})$
  \STATE $e(\Concept{P}) \gets \bar e_{\Concept{P}} / \|\bar e_{\Concept{P}}\|$
  \ENDFOR
  \STATE $\text{ParentEmbs} \gets \{e(\Concept{P})\mid \Concept{P}\in\text{InternalNodes}(T)\}$
  \\[.6\baselineskip]
  \COMMENT{\emph{3. Name parent embeddings by one-to-one matching}}
  \STATE $e(\text{desc}) \gets F(\text{desc})/\|F(\text{desc})\|$ for each $\text{desc}\in\text{Descs}$
  \STATE assignment cost $L_{\text{assign}}\colon (p,\text{desc})\mapsto d(p,e(\text{desc}))$
  \STATE $A \gets \text{LSA}(\text{ParentEmbs},\text{Descs},L_{\text{assign}})$
  \\[.6\baselineskip]
  \RETURN tree $T$ with node embeddings $e(\cdot)$ and parent name assignment $A$
\end{algorithmic}
\end{algorithm}

\subsection{Explainable and Faithful Tree-Traversal Inference with UAES}\label{sec:tree-traversal}
Standard zero-shot classification of a sample $x$ directly picks the leaf class with the most similar embedding to $F(x)$.
Wan et al.~\shortcite{wan2020nbdt} instead suggest to use a hierarchy tree of embeddings for inference via \emph{tree-traversal}:
One hierarchically traverses the tree from root to leaf, at each node picking the child with the closest embedding to $F(x)$. This results in a more explainable classification, since the full path of specialization is available (instead of just \Concept{cat}, \enquote{\Concept{animal} $\to$ \Concept{mammal} $\to$ \Concept{cat}}).
Interestingly, predictions don't necessarily coincide with zero-shot classification:
if a leaf lies at the boundary of its superclass $\Concept{P}$, the centroid of a sibling $\Concept{P}'$ may be closer to $F(x)$, so traversal proceeds into the subtree of $\Concept{P}'$ (cf.\ \cref{fig:tree-traversal}).
From a semantic perspective this can be desirable: assuming interpolatability, a parent covers the area between its children, thereby preserving abstraction for underrepresented superclasses.
\begin{example}
As illustrated in \Cref{fig:tree-traversal}, in a hierarchy of many \Concept{animal} and few \Concept{vehicle} subclasses, tree-traversal can still correctly consider \Concept{coach} a \Concept{vehicle} instead of an outlier animal (\Concept{horse}).
\end{example}

\noindent Nevertheless, a deep traversal into less matching subclasses should be avoided.
Conveniently, the calculated similarities at each node may serve as a certainty score that traversal down that node is reasonable. We suggest to \emph{stop early} at a parent node if none of its children would be sufficiently certain, returning the more reliable parent as prediction instead of a wild guess of a leaf class, as specified in \cref{alg:tree-traversal}.
\begin{example}[UAES]\label{ex:running-cifar10-faithfulness}
In \cref{fig:tree-traversal}, \Concept{vehicle} should be returned when traversal is uncertain between subclasses \Concept{motor vehicle} and \Concept{boat-like}).
Early stopping enables more reliable classification both of unknown classes (\Concept{coach}) as well as known but erroneously classified ones.
\end{example}
\noindent We estimate the cut-off threshold for a node at the $p$th quantile (here: $0.01$) of similarities observed in the training set.
The ideal stopping point is the first node where a \enquote{wrong turn} would be taken---the lowest common ancestor (LCA) of the ground truth and the leaf node predicted by vanilla tree-traversal. The average walking distance of a prediction to that LCA gives an intuitive measure how much reliability early stopping recovers, here abbreviated as last correct node distance.

\begin{algorithm}[tb]
\caption{Tree-traversal with UAES}\label{alg:tree-traversal}
\begin{algorithmic}
\REQUIRE encoder $F$, instance $x$, hierarchy tree $T=(\mathcal{V},E,\Concept{P}_\text{root})$, embeddings $e(\Concept{C})$ for all $\Concept{C}\in\mathcal{V}$, node-wise UAES similarity thresholds $(\tau_{\Concept{P}})_{\Concept{P}\in\mathcal{V}}$, cosine distance $d$
\STATE embedding $v \gets F(x)$
\\[.6\baselineskip]
\COMMENT{\emph{Traverse while the best child remains sufficiently certain}}
\STATE current parent $\Concept{P}\gets \Concept{P}_\text{root}$
\STATE $\text{instanceHierarchy}\gets [\Concept{P}_\text{root}]$
\WHILE{$\Concept{P}$ is no leaf}
\STATE best child $\Concept{C}^* \gets \argmin_{\Concept{C}\in\text{Children}(\Concept{P})} d(v,e(\Concept{C}))$
\STATE best child distance $d_{\text{child}}^* \gets d(v,e(\Concept{C}^*))$
\\[.6\baselineskip]
\COMMENT{\emph{Uncertainty-aware Early Stopping (UAES)}}
\IF{$d_{\text{child}}^* > \tau_{\Concept{P}}$}
\STATE \textbf{break}
\ENDIF
\\[.6\baselineskip]
\COMMENT{Traverse down to best child}
\STATE $\Concept{P}\gets \Concept{C}^*$
\STATE append $\Concept{P}$ to $\text{instanceHierarchy}$
\ENDWHILE

\RETURN prediction $F_{\text{tree-traversed}}(x)=\Concept{P}$, $\text{instanceHierarchy}$
\end{algorithmic}
\end{algorithm}

\paragraph{Measuring Hierarchy Faithfulness.}
While vanilla tree-traversal could theoretically surpass zero-shot performance,
this is not achieved even on simpler embedding spaces \cite{wan2020nbdt}.
Reasons can be: unfavorable multi-modal, scattered or overlapping concept representations in embedding space violating our assumptions \cite{mikriukov2025local}, or unequal covariances ($\approx$ widths) of sibling class clusters as in unbalanced trees. Hence, we
follow the XAI faithfulness notion used by Wan et al.~\shortcite{wan2020nbdt}, which answers how well hierarchical-interpretation predictions coincide with vanilla zero-shot ones.
They operationalize it by interpreting the ratio $\frac{\text{tree inference accuracy}}{\text{zero-shot accuracy}}$ as a proxy measure how \textbf{faithfully} the hierarchy captures the internal similarity structure.
To capture how deeply rooted category confusions are, we suggest \textbf{soft tree inference accuracy} (the average fraction of the correct root-to-leaf path taken during traversal).

\subsection{Verifying Plausibility of Induced Hierarchies}
\label{sec:verify}\label{sec:verification}
Since internal nodes are named and the structure is a tree, extracted hierarchies support both qualitative inspection and quantitative comparison to a reference hierarchy.
We assume a target hierarchy $T'$ represented as a directed acyclic graph, together with a mapping $\mathcal{M}$ from concept-bank entries to nodes in $T'$.
The target $T'$ can be a taxonomic subset of a human-defined ontology $\Ontology$ or the hierarchy of a different encoder.
We introduce two ontology-based verification metrics: (i) the distance to the closest valid tree extracted from the ontology (global fit), and (ii) a hierarchical consistency score $\mathcal{S}_{\text{onto}}$ that checks parent-child edges for matching ontology paths (local fit).
Let $d_{T}$ be the integer walking distance between nodes in the underlying undirected graph of $T$.

\paragraph{Globally Verifying against the Closest Valid Trees.}
We consider a hierarchy plausible with respect to $\Ontology$, if it is a slice thereof.
To quantify how far an extracted hierarchy $T$ is from being plausible, we search for the best matching $T'\subset\Ontology$.

\begin{definition}
Let $d_{\text{tree}}$ be a tree comparison metric \cite{kutsu2010tree}, like Tree Edit Distance \cite{zhang1989simple} or Tree Alignment \cite{jiang1995alignment}.
The \emph{closest valid tree} to $T$ (without renaming)
with respect to ontology $\Ontology{}$ and $d_{\text{tree}}$
is the subtree
\begin{align}
T_{\text{\normalfont valid}} = {\textstyle\argmin_{T'\subset \Ontology}} d_{\text{\normalfont tree}}(T, T')
\quad\text{ s.t. $T'$ is a tree}
\end{align}
\end{definition}
\noindent $T_{\text{\normalfont valid}}$ can serve as a minimal target to optimize for plausibility, and $d_{\text{tree}}(T, T_{\text{valid}})$ as \emph{global} (yet, costly to obtain) plausibility measure.

\paragraph{Hierarchical Consistency Score.}
This metric efficiently locally verifies that parent-child edges in the $T$ respect the hypernymy constraints of the reference ontology $\Ontology$. We ground each parent node $u$ to a representative ontology concept
\begin{equation}
\rho(u) = \text{LCA}_{\Ontology}\left(\{ \mathcal{M}(l) \mid l \in \text{leaves}(u) \}\right).
\end{equation}
For each edge $u \to v$, let $\delta(\rho(u), \rho(v))$ be the shortest directed path length in $\Ontology$.
We define $S_{\text{edge}}(u\to v)=0$ if no path exists, else $S_{\text{edge}}(u\to v)=\min(1, (\gamma \cdot \delta(\rho(u), \rho(v)))^{-1})$. The factor $\gamma=0.5$ is adapted from path-length normalization principles \cite{Wordnet} and allows to tolerate common one-level skips.
We then aggregate over edges $E$ in $T$ to obtain \textbf{Hierarchical Consistency}
\begin{equation}
\mathcal{S}_{\text{onto}}(T) = \frac{1}{|\mathcal{E}|} \sum_{(u\to v)\in \mathcal{E}} S_{\text{edge}}(u\to v).
\end{equation}

\begin{example}[Local Hierarchical Consistency]\label{ex:running-cifar10-verify}
In \cref{fig:example-commitment}, the edge \Concept{animal} $\to$ \Concept{canine} only skips \Concept{mammal} and receives high $S_{\text{edge}}=1$,
while $S_{\text{edge}}(\text{\Concept{canine} $\to$ \Concept{equine}})=0$.
\end{example}

\subsection{Aligning Induced to Target Hierarchies}
\label{sec:align}

The last part of our pipeline aims to post-hoc align the observed similarity structure $T$ of the VLM encoder to a target tree $T'$, like the closest valid tree or a different encoder's semantics.
We here suggest an efficient post-hoc, model-agnostic approach that only learns a transformation for the embedding space. Thereby the loss function accounts for preserving the original similarity structure such that the transformation minimally affects zero-shot performance on the given leaf classes
Ingredients are the tree structure and leaf labels of $T'$, and a training and test split of the leaf concept dataset.
Then, a transformation is obtained in two steps:
\begin{enumerate}[nosep, label=(\arabic*)]
\item Determine suitable \textbf{target latent points} to which the instances should be mapped, in order to fulfill the target similarity structure.
\item Train a \textbf{transformation} of the embedding space, here a 2-layer DNN, on the pairs of original and targeted instance embeddings.
\end{enumerate}
This modularization allows to use efficient manifold learning techniques like UMAP \cite{mcinnes2020umap} in the first step.
Via an optimization procedure
UMAP finds embeddings of a set of original points in a target space, such that the embeddings match targeted pairwise point distances. This is typically used to map points into a \emph{lower}-dimensional space, whilst targeting the pairwise distances from the original space to \emph{preserve} neighborhoods.
We instead map points into the \emph{same} space, and want to \emph{change} neighborhoods to match those of $T'$.
Hence, we initialize the UMAP mapping with the identity, and sum pairwise distances between points $x$, $x'$ of concepts \Concept{C}, \Concept{C}' that ($\alpha_\mathrm{orig}$) anchor to the original cosine geometry, ($\beta_\mathrm{onto}$) drive pairwise embedding distances to align with the class distances in $T'$
, and ($\gamma_\mathrm{midp}$) determine how much the original class distances should be reduced. 
Combining the original cosine geometry and the class distances prevents collapsing class representations, i.e.,
\begin{align}
\SwapAboveDisplaySkip
\begin{aligned}
&d_{\text{emb}}(x, x') =
\overbrace{\alpha_{\textrm{orig}} \cdot d(F(x), F(x'))}^{\mathclap{\text{\scriptsize maintain original instances' dist.}}} \\
&+ \underbrace{\tfrac{1}{Z}\beta_{\textrm{onto}} \cdot d_{T'}(\Concept{C}, \Concept{C}')}_{\text{\scriptsize match ontology class dist.}} -
\underbrace{\gamma_{\textrm{midp}} \cdot d(e(\Concept{C}), e(\Concept{C}'))}_{\text{\scriptsize discard orig. class dist.}}
\end{aligned} 
\label{eq:umap}
\end{align}
\noindent for cosine distance $d$, the undirected tree walking distance $d_{T'}$ in $T'$, the normalization constant \mbox{$Z:=\max d_{T'} / \max_{C_1, C_2} d(e(C_1), e(C_2))$}, and scalar parameters $\alpha_\textrm{orig}$, $\beta_\textrm{onto}$ and $\gamma_\mathrm{midp}$.
It is a combination of maintaining as much as possible of the \emph{original distances} between instances within a class representation,
whilst driving the embedding distances to match the length of the \emph{shortest path between the classes} in $T'$.
During training, the second term is normalized by the maximum path distance and maximum class embedding distance in the dataset.
Given these target distances $d_{\text{emb}}(x,x')$, UMAP creates a dataset of pairs each mapping an original embedding of a probing sample to a transformed counterpart embedding vector. The pairwise distances between the transformed embeddings are encouraged to match the target distances.
Since the mapping is only valid on the probing samples, the second step then trains a continuous function to extend this mapping to the complete embedding space. This can be applied before zero-shot classification on the latent space to align representations closer with a target class hierarchy.

\section{Experiments}\label{sec:experiments}
We conduct an empirical study to investigate faithfulness and utility of the extracted hierarchies for tree-traversal (%
\cref{sec:downstream}); compare plausibility of hierarchies across modalities and models (
\cref{sec:plausibility}), and to measure how easily embedding spaces can be post-hoc aligned to target hierarchies (%
\cref{sec:exp-alignment}).

\textit{The supplementary provides further details on the setup, ablation study, tables with numerical results of all plots, hypothesis tests (p-values are $<10^{-5}$ if not reported).}

\subsection{Experimental Settings}

A hyperparameter study selected WordNet-based over Llama3-based or combined descriptions for generating concept bank embeddings as optimal (see supplementary). WordNet is used for the general, WordNet+Llama3 for the specialized CUB dataset.
The following settings apply to the total of 156 main experiments.


\paragraph{Models.}
We consider in total 13 pre-trained foundational VLMs from 4 families
\cite{miller2024open}: 
\textbf{CLIP} (C) with 9 different backbones \cite{radford2021learning},
%
\textbf{ALIGN} \cite{jia2021Scalingvisual}, 
%
\textbf{FLAVA} \cite{singh2022FLAVAFoundational}, 
%
and \textbf{SigLIP} (v2) with 224px and 384px backbones \cite{zhai2023sigmoid}.
%

\paragraph{Leaf Class Datasets.}
We use CIFAR-10/100 \cite{alex2009learning} and ImageNet \cite{deng2009imagenet} (10/100/1000 leaves) as standard datasets of increasing hierarchy complexity, and CUB \cite{wah2011caltechucsd} as a specialized 200-class bird dataset. The train split is used for hierarchy extraction and alignment training; the test split is used to assess faithfulness and plausibility.

\paragraph{Ontologies and Concept Bank.}
We use the WordNet lexical database \cite{fellbaum2010wordnet} as a concept bank and SUMO \cite{niles2001standard}, OpenCyc \cite{Matuszek}, and Yago \cite{YAGO3AK} as reference general domain ontologies, linked via existing mappings. We include nouns and adjectives, and treat \Concept{Subclass} and \Concept{InstanceOf} as parent relations.

\paragraph{Metrics.}
We measure \textbf{discriminability} as zero-shot performance (ZS, $\uparrow$), \textbf{faithfulness} via our faithfulness ratio, (soft) tree-traversal inference (tree / tree soft, $\uparrow$) and distance to the last correct node ($\downarrow$), \textbf{plausibility} using our hierarchical consistency score ($\uparrow$), and normalized constrained Unordered Tree Edit Distance (nUTED, $\downarrow$) \cite{zhang1996constrained} to obtain the closest valid tree. The latter is the minimal number of deletion/insertion/leaf rename operations to turn one tree into the other, normalized to $[0,1]$ by the operation number of the trivial delete-all-insert-all solution \cite{RICOJUAN20031417}. Two randomly constructed binary trees for 1000 leaf nodes have a $\text{nUTED}$ of approximately $0.54$ (see supplementary).

\subsection{Explaining Hierarchies: Faithfulness}
\label{sec:faithfulness}\label{sec:downstream}
A qualitative global explanation of VLM hierarchies is demonstrated in \cref{fig:example-commitment}.
To quantitatively assess explanation quality in terms of faithfulness, we use the (relative) performance of local explanations via tree-traversal inference as a proxy.
We conducted an ablation study across different
leaf set complexities,
VLM model architectures,
and the chosen encoder (text, image, mean of both).
%

\paragraph{More Leaves Diminish Faithfulness.}
\begin{table}
\centering
\sisetup{table-format=2.1(1), table-align-uncertainty=false}
\begin{footnotesize}
\begin{tabular}{@{}l*{4}{@{\,}S}@{}}
  \toprule
  {} & {tree ($\uparrow$)} & {tree soft ($\uparrow$)} & {faithfulness ($\uparrow$)} & {ZS ($\uparrow$)} \\\midrule
  CIFAR10 & 81.8341 \pm 11.4794 & 88.9553 \pm 8.31674 & 94.2477 \pm 6.75924 & 86.5872 \pm 8.63874 \\
  CIFAR100 & 35.4682 \pm 14.2324 & 59.4992 \pm 11.9802 & 54.536 \pm 15.4268 & 63.4677 \pm 13.3033 \\
  CUB & 21.6555 \pm 17.9089 & 35.5568 \pm 25.6268 & 37.2673 \pm 15.5722 & 49.422 \pm 33.1719 \\
  ImageNet & 18.0497 \pm 11.2519 & 38.9541 \pm 17.7296 & 26.0853 \pm 15.255 & 67.5365 \pm 8.7422 \\
  \bottomrule
\end{tabular}
\end{footnotesize}
\caption{Faithfulness, tree-traversal accuracy (tree), soft tree-traversal accuracy (tree soft), and zero-shot performance (ZS) by dataset, averaged over models and encoders.}\label{fig:faithfulness_by_dataset}
\end{table}

The dominating effect is that a higher number of leaf nodes considerably decreases faithfulness as shown in \cref{fig:faithfulness_by_dataset}: average faithfulness drops from $\num{94.2\pm 6.759}$ on CIFAR-10 (10 classes) to $\num{26.1\pm 15.26}$
on ImageNet (1000 classes). Similar decrease in soft tree-traversal performance ($89$ to $38$) indicates that tree-traversal inference takes a wrong turn more easily in deeper trees.
While the effect is smaller on image-embedded leaves, it is still significant.
As a first pathway, this motivates use of our suggested early stopping.

\paragraph{Early Stopping Improves Hierarchical Inference.}

We evaluate whether uncertainty-aware early stopping (UAES) can correctly capture wrong turns in hierarchical inference.
\Cref{tab:last-correct-node-detection} reports the undirected tree distance between the predicted output (leaf or early-stopped internal node) and the last correct node, defined as the LCA of the ground-truth leaf and the leaf predicted by vanilla traversal.
Early stopping reduces this distance on the larger label sets (CIFAR100/CUB/ImageNet), which means that it often returns semantically appropriate super-categories instead of committing to an incorrect leaf; on CIFAR10 the effect is negligible.
Notably, measuring the distance to the ground truth node not in the tree but in the reference ontology SUMO has no significant effect. This validates that early stopping does not change the quality of the semantic alignment against a target ontology, which we inspect in detail next.

\begin{table}
\centering
\footnotesize
\resizebox{.55\linewidth}{!}{%
\begin{tabular}{@{}lSS@{}}
  \toprule
  & {early stop (ours)}            & {vanilla}                       \\
  \midrule
  CIFAR10  & 0.348577 \pm 0.173017          & \bfseries 0.328772 \pm 0.197652 \\
  CIFAR100 & \bfseries 2.1177 \pm 0.560399  & 2.42832 \pm 0.586459            \\
  CUB      & \bfseries 1.11294 \pm 0.768076 & 1.66217 \pm 0.596943            \\
  ImageNet & \bfseries 2.47399 \pm 0.980289 & 4.27876 \pm 0.996317            \\
  \bottomrule
\end{tabular}}~%
\includegraphics[width=.45\linewidth,align=c]{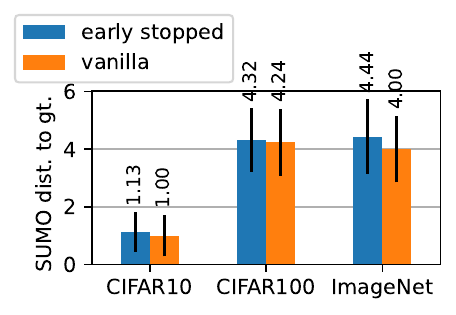}
\caption{Tree-traversal early stopping returns internal nodes instead of incorrect leaves.
Left: raw last correct node distance ($\downarrow$), averaged over encoders and models.
Right: average semantic distance ($\downarrow$) from the predicted node to the ground-truth concept in the SUMO ontology (walking distance).
Best per dataset in bold.}\label{tab:last-correct-node-detection}
\end{table}

\subsection{Agreement with Human Ontologies}
\label{sec:plausibility}
We quantify plausibility by comparing extracted hierarchies against human reference ontologies (OpenCyc, SUMO, and Yago), using the hierarchical consistency metric introduced in \cref{sec:verification}.
Our evaluation is driven along our desiderata:
Can discriminability and plausibility be achieved simultaneously?
And does the text-to-image alignment in CLIP training also induce semantically aligned hierarchies?

\begin{figure}[t]
\raggedleft
\includegraphics[width=\linewidth]{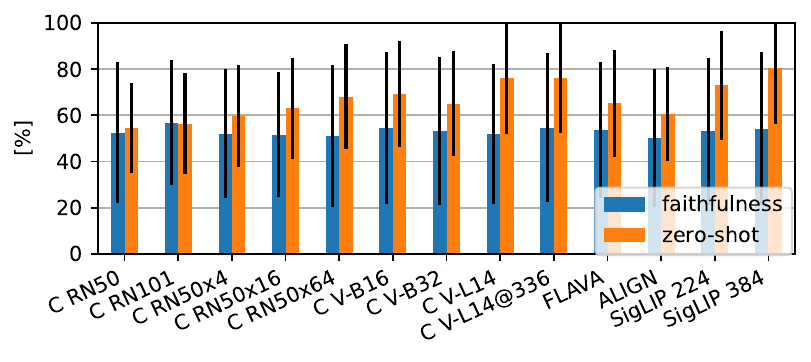}%
\\[-2.9\baselineskip]
\includegraphics[width=\linewidth]{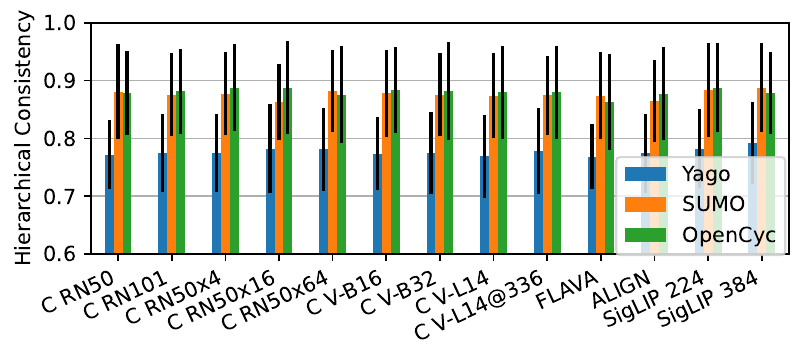}%
\\[-\baselineskip]
\caption{Faithfulness (top) versus plausibility (bottom) results for different models and backbones,
averaged over 3 datasets and 3 leaf encoding types (image, text, both).}
\label{fig:faithfulness-by-model}
\end{figure}

\begin{figure*}[t]
\begin{subfigure}[b]{.62\linewidth}
\includegraphics[width=\linewidth, align=t]{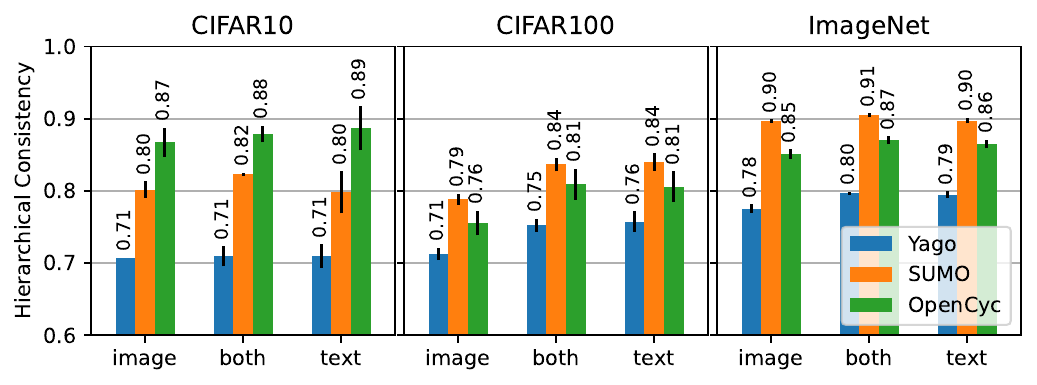}%
\caption{%
  Hierarchical consistency ($\uparrow$) against OpenCyc, SUMO, and Yago across CIFAR10/100 and ImageNet.
}
\label{fig:faithfulness-by-leaf-mode-a}
\end{subfigure}%
\hfill
\begin{subfigure}[b]{.35\linewidth}
\small\centering
\sisetup{table-format=2.0(2), table-align-uncertainty=false}%
\begin{tabular}[t]{@{}l SSS@{}}
  \toprule
  {} & {tree} & {faithful} & {zero-shot} \\\midrule
  image & \bfseries 48.736 \pm 23.5249    & \bfseries 63.1772 \pm 21.941 & \bfseries 74.4585 \pm 12.0352 \\
  both & 44.476 \pm 27.1792                & 57.3509 \pm 27.7561 & 74.1182 \pm 12.1082 \\
  text & 24.5437 \pm 30.4675              & 38.574 \pm 31.9785 & 51.6833 \pm 31.0595 \\
  \bottomrule
\end{tabular}
\\\strut
\caption{%
Faithfulness metrics averaged over model, dataset, and encoder.}\label{fig:faithfulness-by-leaf-mode-b}
\end{subfigure}
\caption{Effect of leaf embedding modality: Results of verifying and comparing encoders against a target ontology (\cref{fig:faithfulness-by-leaf-mode-a,fig:faithfulness-by-leaf-mode-b}).
For results on comparing encoders against each other see \cref{fig:uted}.}
\label{fig:faithfulness-by-leaf-mode}
\end{figure*}

\begin{figure}
\centering
\includegraphics[width=\linewidth, align=t]{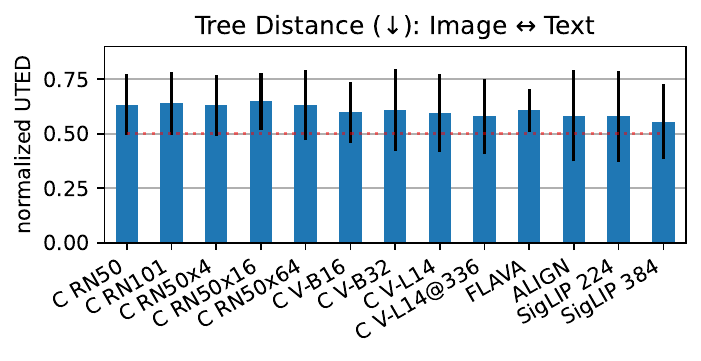}
\caption{nUTED ($\downarrow$) between trees induced by \textbf{image} and \textbf{text} encoders of each the same VLM;
averaged over standard datasets. nUTED$>0.5$ marks worse than random.}
\label{fig:uted}
\end{figure}

\paragraph{Plausibility--Discriminability Trade-off.}
When considering per-model metrics, zero-shot performance has little predictive value about the plausibility, as visualized in \cref{fig:faithfulness-by-model}.
More interestingly zero-shot performance and faithfulness are both significantly negatively correlated with hierarchical consistency across all ontologies (for zero-shot and SUMO/OpenCyc/Yago: p-values $<0.006$ and Pearson r $<-0.45$/$-0.21$/$-0.29$; for faithfulness: Pearson r$<-0.62$/$-0.18$/$-0.61$ and p-value $<0.023$ for OpenCyc, else $0$).
Hence, \textbf{there seems to be a consistent trade-off between
discriminability
of leaf classes, and plausibility of the induced semantic structure.}

\paragraph{Modality Gap: Text versus Image Hierarchies.}
\label{sec:modality}

As shown in \cref{fig:faithfulness-by-leaf-mode},
image leaf embeddings achieve significantly higher zero-shot and faithfulness scores compared to text leaf embeddings ($+\num{22.244\pm 13.486}$ percent points), which get instead consistently higher plausibility values ($+\num{0.015\pm 0.032}$, p-value$<0.002$), see \cref{fig:faithfulness-by-leaf-mode}.
This is confirmed by the significantly different trees, measured via average nUTED between trees when switching from image to text leaf embeddings (see \cref{fig:uted}).
Combining text and image information when embedding leaves offers a useful trade-off between discriminability and plausibility (cf.\ \cref{fig:example-commitment} and \cref{fig:faithfulness-by-leaf-mode}).

We then ask whether these trends hold consistently across model families and backbones.

\paragraph{Across-Model Comparison.}
\label{sec:comparison}
Across VLM families, we observe substantial variation in both faithfulness and ontological plausibility.
\Cref{fig:faithfulness-by-model} summarizes these trends and highlights that higher zero-shot accuracy does not necessarily imply a hierarchy that is more consistent with human ontologies. 

In the next section we evaluate in how far our alignment procedure also confirms this trade-off between faithfulness and plausibility.

\subsection{Post-hoc Alignment to Target Hierarchies} 
\label{sec:exp-alignment}
We here investigate whether learned ontologies can successfully be post-hoc corrected, without destroying zero-shot accuracy.
For this we consider the following tasks:
\begin{itemize}
\item Validation: Swap two leaves in the semantic tree.
\item Modality alignment: Turn a tree for image leaf embeddings into the corresponding one for text embeddings.
\item Ontological commitment: Steer to the closest valid tree.
\end{itemize}
We measure proximity of a current to a target ontology by turning them into their distance matrices and taking the Frobenius norm of their difference. $\Delta\Ontology$ is defined as the relative change in distance to the target from after manipulation to before. $\text{ZS}_{\makebox[0pt][l]{\text{\scriptsize text}}}^{\text{\scriptsize Orig}}$, $\text{ZS}_{\makebox[0pt][l]{\text{\scriptsize midp}}}^{\text{\scriptsize Orig}}$, $\text{ZS}_{\makebox[0pt][l]{\text{\scriptsize text}}}^{\text{\scriptsize UMAP}}$ and $\text{ZS}_{\makebox[0pt][l]{\text{\scriptsize midp}}}^{\text{\scriptsize UMAP}}$ denote zero-shot accuracy before (Orig) and after manipulation (UMAP) and base on (transformed) text embeddings and class midpoints.
For this study, we investigate four models, ALIGN, CLIP with ViT-L14@336px backbone, SigLIP-224 and FLAVA (as specified before), on the validation task and run experiments on $2,500$ random training images and evaluate on $10,000$ test images of the CIFAR10 dataset.
Averaged over three seeds, 10 random swaps and all four models, according to our hyperparameter grid-search the parameter setting $\alpha_{\textrm{orig}}=2.0$, $\beta_{\textrm{onto}}=1.0$,
$\gamma_{\textrm{midp}}=2.0$, $\textrm{N}_\textrm{UMAP} = 250$, minimizes a combined evaluation score $\Delta\Ontology-\text{ZS}_{\makebox[0pt][l]{\text{\scriptsize midp}}}^{\text{\scriptsize UMAP}}$, which balances ontology alignment and zero-shot performance.


Results for the best parameter setting are presented in \cref{tab:ontology-manipulation}. The general finding is that our approach can cheaply initialize transformations from one latent ontology to another, being capable to
fully align the original clip embedding with an embedding where automotive and frog have been swapped.
There is, however, a trade-off between zero-shot accuracy and successful ontology matching, which reveals how much alignment and class discriminability compete. A comparison against a linear transformation baseline shows how far the true transformation is from being (simply) linear (see supplementary). Results for steering of class hierarchies towards their text embeddings and closest valid trees confirm our findings and can be found in the supplementary as well as the details on the hyperparameter study.

\begin{table*}
\centering
\sisetup{table-auto-round,table-format=1.2(2), uncertainty-mode=separate}
\footnotesize
\begin{tabular}{@{}lrrrrrr@{}}
\toprule
\makecell{ \\ Model} & \makecell{$\Delta\Ontology$ \\ $\downarrow$} & \makecell{nUTED \\ $\downarrow$} & \makecell{{$\text{ZS}_{\makebox[0pt][l]{\text{\scriptsize text}}}^{\text{\scriptsize UMAP}}$}\\{$\uparrow$}} & \makecell{{$\text{ZS}_{\makebox[0pt][l]{\text{\scriptsize midp}}}^{\text{\scriptsize UMAP}}$}\\$\uparrow$} & \makecell{{$\text{ZS}_{\makebox[0pt][l]{\text{\scriptsize text}}}^{\text{\scriptsize Orig}}$}\\$\uparrow$} & \makecell{{$\text{ZS}_{\makebox[0pt][l]{\text{\scriptsize midp}}}^{\text{\scriptsize Orig}}$}\\{$\uparrow$}} \\
\midrule
\multicolumn{7}{c}{ $\alpha_{\textrm{orig}}=2.0$, $\beta_{\textrm{onto}}=1.0$, $\gamma_{\textrm{midp}}=2.0$, $\textrm{N}_\textrm{UMAP} = 250$} \\
\midrule
ALIGN & 0.589 $\pm$ 2.9e-01 & 0.080 $\pm$ 3.7e-03 & 0.579 $\pm$ 1.3e-03 & 0.701  & 0.753  & 0.804 \\
C V-L14@336  & 0.850 $\pm$ 2.8e-01 & 0.098 \textcolor{gray}{$\pm$ 7.1e-04} & 0.732 $\pm$ 4.5e-03 & 0.858  & 0.946  & 0.958  \\
FLAVA & 0.731 $\pm$ 3.9e-01 & 0.072 $\pm$ 2.2e-03 & 0.688 $\pm$ 1.1e-03 & 0.786 & 0.890  & 0.923  \\
SIGLIP 224 & 0.720 $\pm$ 8.6e-02 & 0.102 \textcolor{gray}{$\pm$ 3.7e-04} & 0.723 $\pm$ 2.4e-03 & 0.837  & 0.914  & 0.931  \\
\bottomrule
\end{tabular}
\caption{Results of the ontology manipulation for two swapped leaves; averaged over 10 random swaps and 3 seeds. We report variances exceeding $1.0e-03$ (see supplementary for all variances).}
\label{tab:ontology-manipulation}
\end{table*}

\section{Discussion and Outlook}
Our experiments revealed a remaining \textbf{faithfulness gap} of our explainable surrogates to the true model behavior.
Future work should close this gap further, e.g., by
shifting to \textit{distribution}-based \cite{mikriukov2025local,lee2025conceptbased}
and \textit{location}-aware \cite{fong2018net2vec} concept representations, and
removing any \textit{background bias} in the probing images \cite{schwalbe2025background}.
Nevertheless, it should be noted that even imperfect XAI surrogates remain useful and are widely used for knowledge and test-case discovery \cite{ribeiro2016why}.

Interesting extensions of our technique
would be \emph{concept combinations} like \Concept{black cat}, e.g., using basis decomposition \cite{zhang2021invertible};
towards arbitrary vision \emph{models}, e.g., using locked-image text approaches \cite{zhai};
comparing against specialized ontologies;
more types of \emph{assumptions} \cite{speer2017conceptnet}
and arbitrary 
\emph{graph} structures;
and \emph{ante-hoc} ontology insertion during training.

\textbf{Beyond applications to bias removal and model selection}, we expect that our technique can \emph{guide specialization of VLMs} using domain specific ontologies,
including search for appropriate context prompts for a specific domain.
Furthermore, relaxing the strong explainability assumptions on concept distributions in latent space could not only \emph{inform concept entanglement analysis}, but also improve UAES
and thus semantic uncertainty-informed VLM zero-shot inference.
Further applications could invert the here proposed analysis and use a VLM to validate human ontologies: E.g.,
\emph{comparing human-defined ontologies} with respect to their coverage of VLM knowledge, and \emph{expanding or correcting human-defined ontologies} from DNN-learned knowledge.


\section{Conclusion}

This paper presented a principled approach to extract the learned class hierarchy induced by a given set of leaf classes from a VLM.
Furthermore, we introduced techniques to measure and improve faithfulness and plausibility, where the latter makes direct use of existing human-defined knowledge bases.
We generally could uncover a
mismatch
between class hierarchies learned by the text and by the image encoders.
Our suggested method to post-hoc train ontology manipulating transformations was able to improve the matching of a learned ontology towards a desired one, with only moderate loss on zero-shot accuracy.
This suggests that human-interpretable knowledge can directly be used to align VLM representations with human expectations, and potentially help in closing the gap in text-to-image alignment.

\appendix

\section*{Ethical Statement}

There are no ethical issues.

\section*{Acknowledgments}
G.S.\ and S.T.\ acknowledge support through the project ``chAI'' funded by the German Federal Ministry of Research, Technology and Space (BMFTR), grant no.\ 16IS24058.
A.M.\ and M.R.\ gratefully acknowledge financial support from the BMFTR within the project ``REFRAME'' (grant no.\ 16IS24073C). E.H., A.M.\ and M.R.\ also gratefully acknowledge support through the BMFTR junior research group ``UnrEAL'' (grant no.\ 16IS22069).
J.H.L.\ was supported by the German Research Foundation (DFG), project no.\ 551629603. M.K.\ acknowledges support through the project "NXT GEN AI METHODS – Generative Methoden für Perzeption, Prädiktion und Planung" funded by the German Federal Ministry for Economic Affairs and Energy.
The authors are solely responsible for the content of this publication.

\section*{Contribution Statement}
Authors Jae Hee Lee and Matthias Rottman contributed equally in advising.

\bibliographystyle{named}
\bibliography{main}

\newpage
\onecolumn
\doparttoc 
\faketableofcontents 
\renewcommand \thepart{}
\renewcommand \partname{}
\appendix
\addcontentsline{toc}{section}{Appendix} 
\part{Appendix with Supplementary Material} 
\parttoc 

\input{appendix}

\end{document}

%% file: appendix.tex
\section{Tabular Results for Plots in the Paper}  

Tabular results for the plots in the main paper can be found in (in order of the paper) in:
\begin{itemize}
  \item \cref{tab:dist_to_gt_in_onto_by_early_stop}: Comparison of tree-traversal with and without uncertainty-aware early stopping (UAES)
  \item \cref{tab:faithfulness_by_model-onto_m2_by_model}: Faithfulness and plausibility by model
  \item \cref{tab:onto_m2_by_leaf_source_by_dataset}: Faithfulness and plausibility by encoder type and dataset
  \item \cref{tab:normalized_uted_by_model}: nUTED between text and image encoders
\end{itemize}

\begin{table}[b]
  \centering
  \begin{tabular}{lSS}
    \toprule
    {dataset} & {early stopped} & {vanilla} \\\midrule
    CIFAR10 & 1.12783 \pm 0.685725 & 1.00161 \pm 0.722408 \\
    CIFAR100 & 4.31606 \pm 1.09736 & 4.24135 \pm 1.16467 \\
    ImageNet & 4.43816 \pm 1.28787 & 4.00128 \pm 1.15433 \\
    \bottomrule
  \end{tabular}
  \caption{Walking distance of early-stopped versus vanilla tree-traversal predictions to ground truth node in SUMO.}\label{tab:dist_to_gt_in_onto_by_early_stop}
\end{table}

\begin{table}[b]
  \footnotesize\centering
  \resizebox{\linewidth}{!}{%
    \begin{tabular}{lSSSS|SSS}
      \toprule
      & \multicolumn{3}{c}{Faithfulness} & \multicolumn{3}{c}{Hierarchical Consistency} \\
      {model} & {tree} & {tree soft} & {faithfulness} & {zero-shot} & {Hier. Consist. (Yago)} & {Hier. Consist. (SUMO)} & {Hier. Consist. (OpenCyc)} \\\midrule
      C RN50 & 31.8317 \pm 25.8721 & 51.9281 \pm 25.9067 & 52.4381 \pm 30.5322 & 54.511 \pm 19.6896 & 0.772002 \pm 0.060379 & 0.880986 \pm 0.0818652 & 0.878441 \pm 0.0725994 \\
      C RN101 & 34.586 \pm 25.8257 & 56.7676 \pm 24.9015 & \bfseries 56.822 \pm 27.1548 & 56.3955 \pm 21.7153 & 0.775045 \pm 0.0677054 & 0.87599 \pm 0.0717954 & 0.881397 \pm 0.0742818 \\
      C RN50x4 & 33.7712 \pm 25.9481 & 52.8612 \pm 25.6748 & 52.1117 \pm 28.0943 & 59.6769 \pm 21.8947 & 0.774953 \pm 0.0672519 & 0.877218 \pm 0.0718691 & \bfseries 0.888085 \pm 0.0748344 \\
      C RN50x16 & 35.4989 \pm 25.0989 & 53.7402 \pm 24.8214 & 51.6501 \pm 27.2003 & 63.1026 \pm 21.8353 & 0.782073 \pm 0.0771141 & 0.863158 \pm 0.0652447 & 0.887669 \pm 0.0807843 \\
      C RN50x64 & 37.9839 \pm 29.3122 & 55.7146 \pm 26.9568 & 51.2353 \pm 30.7231 & 68.1286 \pm 22.5873 & 0.781446 \pm 0.0720746 & 0.882568 \pm 0.0713157 & 0.875987 \pm 0.0834133 \\
      C V-B16 & 42.4239 \pm 32.7171 & 58.4968 \pm 29.0249 & 54.5209 \pm 32.7708 & 69.3772 \pm 22.9113 & 0.773322 \pm 0.0632725 & 0.877897 \pm 0.0757745 & 0.883608 \pm 0.0742335 \\
      C V-B32 & 39.3589 \pm 32.1654 & 56.348 \pm 30.3209 & 53.113 \pm 32.0678 & 65.0818 \pm 22.5674 & 0.774371 \pm 0.0708237 & 0.875428 \pm 0.0718753 & 0.882235 \pm 0.0853795 \\
      C V-L14 & 43.7714 \pm 32.2428 & 57.7926 \pm 28.9383 & 51.8084 \pm 30.2905 & 76.0685 \pm 24.1842 & 0.768809 \pm 0.0722783 & 0.874007 \pm 0.0741507 & 0.879998 \pm 0.0802813 \\
      C V-L14@336 & 45.9705 \pm 33.7939 & \bfseries 60.4939 \pm 29.9722 & 54.6918 \pm 32.1004 & 76.3504 \pm 23.8457 & 0.778823 \pm 0.0744674 & 0.87457 \pm 0.0689306 & 0.879988 \pm 0.0804804 \\
      FLAVA & 39.8913 \pm 31.1669 & 56.0276 \pm 27.8082 & 53.674 \pm 29.362 & 65.197 \pm 23.1655 & 0.768522 \pm 0.0562115 & 0.874229 \pm 0.0759015 & 0.863401 \pm 0.082934 \\
      ALIGN & 33.8733 \pm 26.0061 & 50.3529 \pm 27.1703 & 50.0976 \pm 29.9542 & 60.5881 \pm 20.2608 & 0.774643 \pm 0.0683485 & 0.865078 \pm 0.070666 & 0.877738 \pm 0.0809926 \\
      SigLIP 224 & 43.0736 \pm 31.4987 & 55.568 \pm 31.7331 & 52.9917 \pm 31.8282 & 72.9817 \pm 23.509 & 0.781961 \pm 0.0685354 & 0.88368 \pm 0.0819326 & 0.887669 \pm 0.0770586 \\
      SigLIP 384 & \bfseries 48.2399 \pm 33.0735 & 58.546 \pm 32.8067 & 54.2882 \pm 33.1886 & \bfseries 80.3341 \pm 24.0089 & \bfseries 0.792543 \pm 0.0708036 & \bfseries 0.887463 \pm 0.077087 & 0.879183 \pm 0.0709695 \\
      \bottomrule
    \end{tabular}
  }
  \caption{Faithfulness and plausibility metrics by model averaged over encoders.}\label{tab:faithfulness_by_model-onto_m2_by_model}
\end{table}

\begin{table*}
  \centering
  \sisetup{table-format=2.1(2)}
  \footnotesize
  \begin{tabular}{llSSS|*{3}{S[table-format=3.2(1)]}}
    \toprule
    {} & {} & {tree} & {faithfulness} & {zero-shot} & {Yago} & {SUMO} & {OpenCyc} \\\midrule[\heavyrulewidth]
    \multirow[c]{3}{*}{CIFAR10} & image & 85.3085 \pm 8.75938 & 96.8987 \pm 2.29032 & 87.8992 \pm 7.16286 & 0.707672 \pm 0.0 & 0.801913 \pm 0.0116783 & 0.86735 \pm 0.0203476 \\
    & both & 86.6092 \pm 7.63824 & 97.7307 \pm 1.12502 & 88.5554 \pm 6.90112 & 0.709402 \pm 0.0139087 & 0.822792 \pm 0.00243627 & 0.878632 \pm 0.0114746 \\
    & text & 73.5846 \pm 13.0658 & 88.1136 \pm 8.75481 & 83.3069 \pm 10.9394 & 0.710063 \pm 0.01623 & 0.798074 \pm 0.029357 & 0.886859 \pm 0.0307789 \\
    \midrule
    \multirow[c]{3}{*}{CIFAR100} & image & 40.6992 \pm 9.92201 & 61.2242 \pm 7.31737 & 66.1015 \pm 11.3863 & 0.712712 \pm 0.00871275 & 0.788141 \pm 0.0076518 & 0.75529 \pm 0.017071 \\
    & both & 45.3423 \pm 10.3877 & 66.3576 \pm 7.21879 & 67.9485 \pm 11.1241 & 0.752352 \pm 0.00917003 & 0.836604 \pm 0.00943767 & 0.80934 \pm 0.0213515 \\
    & text & 20.3631 \pm 7.25594 & 36.0261 \pm 8.77176 & 56.3531 \pm 14.9311 & 0.757353 \pm 0.0141677 & 0.839798 \pm 0.0122311 & 0.805894 \pm 0.0215226 \\
    \midrule
    \multirow[c]{3}{*}{CUB} & image & 42.5878 \pm 7.4302 & 56.1925 \pm 4.49686 & 75.3631 \pm 8.0489 & 0.871241 \pm 0.0245002 & 0.970921 \pm 0.0337067 & 0.99485 \pm 0.00980313 \\
    & both & 21.4068 \pm 4.70454 & 31.096 \pm 4.81788 & 68.5484 \pm 9.61554 & 0.909653 \pm 0.0227724 & 0.976515 \pm 0.0314937 & 1.0 \pm 0.0 \\
    & text & 0.971828 \pm 0.300213 & 24.5134 \pm 10.9077 & 4.35464 \pm 1.27898 & 0.824238 \pm 0.0468371 & 0.981673 \pm 0.0192039 & 0.98052 \pm 0.0202277 \\
    \midrule
    \multirow[c]{3}{*}{ImageNet} & image & 26.3483 \pm 4.32965 & 38.3935 \pm 3.55872 & 68.4702 \pm 7.66527 & 0.775304 \pm 0.00587314 & 0.896837 \pm 0.00227036 & 0.851091 \pm 0.00654807 \\
    & both & 24.5455 \pm 4.50817 & 34.2193 \pm 4.11169 & 71.4206 \pm 7.12255 & 0.796978 \pm 0.00199839 & 0.905278 \pm 0.00238533 & 0.870283 \pm 0.00541176 \\
    & text & 3.25538 \pm 2.1025 & 5.643 \pm 4.34291 & 62.7186 \pm 9.49182 & 0.794736 \pm 0.00482225 & 0.897398 \pm 0.00394712 & 0.864874 \pm 0.00611075 \\
    \bottomrule
  \end{tabular}
  \caption{Plausibility (Hierarchical Consistency) and faithfulness metrics per leaf encoding strategy and per dataset, averaged over models.}\label{tab:onto_m2_by_leaf_source_by_dataset}
\end{table*}

\begin{table}
  \centering
  \begin{tabular}{lS}
    \toprule
    {model} & {nUTED} \\\midrule
    clip RN50 & 0.634006 \pm 0.140378 \\
    clip RN101 & 0.639604 \pm 0.145299 \\
    clip RN50x4 & 0.630569 \pm 0.138743 \\
    clip RN50x16 & 0.647948 \pm 0.131248 \\
    clip RN50x64 & 0.632416 \pm 0.162033 \\
    clip ViT-B/16 & 0.59854 \pm 0.138759 \\
    clip ViT-B/32 & 0.610196 \pm 0.187879 \\
    clip ViT-L/14 & 0.59429 \pm 0.177569 \\
    clip ViT-L/14@336px & 0.579882 \pm 0.172795 \\
    flava flava-full & 0.607504 \pm 0.0972459 \\
    align align-base & 0.58204 \pm 0.208859 \\
    siglip 224 & 0.58035 \pm 0.209051 \\
    siglip 384 & 0.555326 \pm 0.172052 \\
    \bottomrule
  \end{tabular}
  \caption{nUTED between each text and image encoder of the same VLM, averaged over datasets.}\label{tab:normalized_uted_by_model}
\end{table}

\begin{table}
  \centering
  \begin{tabular}{lSSSS}
    \toprule
    {} & {tree inference acc.} & {tree inference soft acc.} & {faithfulness} & {zero-shot acc.} \\\midrule
    CIFAR10 & 81.8341 \pm 11.4794 & 88.9553 \pm 8.31674 & 94.2477 \pm 6.75924 & 86.5872 \pm 8.63874 \\
    CIFAR100 & 35.4682 \pm 14.2324 & 59.4992 \pm 11.9802 & 54.536 \pm 15.4268 & 63.4677 \pm 13.3033 \\
    CUB & 21.6555 \pm 17.9089 & 35.5568 \pm 25.6268 & 37.2673 \pm 15.5722 & 49.422 \pm 33.1719 \\
    ImageNet & 18.0497 \pm 11.2519 & 38.9541 \pm 17.7296 & 26.0853 \pm 15.255 & 67.5365 \pm 8.7422 \\
    \bottomrule
  \end{tabular}
  \includegraphics[width=.4\linewidth]{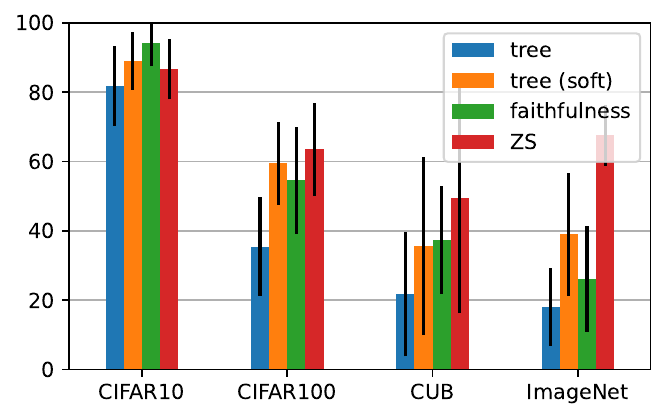}
  \caption{Faithfulness metrics by dataset in percent, averaged over models and encoders.}\label{tab:faithfulness_by_dataset}\label{fig:faithfulness-by-dataset}
\end{table}

\section{Faithfulness Results: More Detailed Comparisons}  
This chapter provides additional, more finegrained statistical insights into the results summarized in the main paper.

\begin{itemize}
  \item \cref{tab:faithfulness_by_dataset}: An additional summary of the faithfulness metrics per dataset
\end{itemize}

\begin{table*}
  \centering
  \includegraphics[width=.8\linewidth]{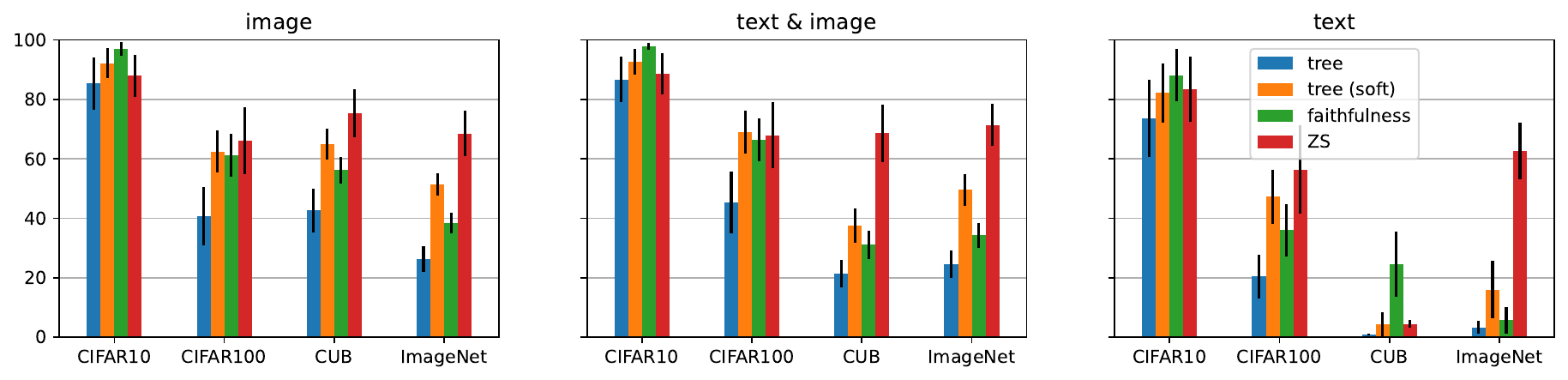}\\%
  \begin{tabular}{llSSSS}
    \toprule
    {encoder} & {dataset} & {tree} & {tree soft} & {faithfulness} & {zero-shot} \\
    \midrule
    \multirow[c]{4}{*}{image} & CIFAR10 & 85.3085 \pm 8.75938 & 92.1122 \pm 5.09308 & 96.8987 \pm 2.29032 & 87.8992 \pm 7.16286 \\
    & CIFAR100 & 40.6992 \pm 9.92201 & 62.4023 \pm 7.07137 & 61.2242 \pm 7.31737 & 66.1015 \pm 11.3863 \\
    & CUB & 42.5878 \pm 7.4302 & 64.9325 \pm 5.32752 & 56.1925 \pm 4.49686 & 75.3631 \pm 8.0489 \\
    & ImageNet & 26.3483 \pm 4.32965 & 51.2988 \pm 3.70023 & 38.3935 \pm 3.55872 & 68.4702 \pm 7.66527 \\
    \multirow[c]{4}{*}{text \& image} & CIFAR10 & 86.6092 \pm 7.63824 & 92.6359 \pm 4.22564 & 97.7307 \pm 1.12502 & 88.5554 \pm 6.90112 \\
    & CIFAR100 & 45.3423 \pm 10.3877 & 68.9042 \pm 7.26669 & 66.3576 \pm 7.21879 & 67.9485 \pm 11.1241 \\
    & CUB & 21.4068 \pm 4.70454 & 37.4952 \pm 5.78489 & 31.096 \pm 4.81788 & 68.5484 \pm 9.61554 \\
    & ImageNet & 24.5455 \pm 4.50817 & 49.5654 \pm 5.29652 & 34.2193 \pm 4.11169 & 71.4206 \pm 7.12255 \\
    \multirow[c]{4}{*}{text} & CIFAR10 & 73.5846 \pm 13.0658 & 82.1177 \pm 9.95571 & 88.1136 \pm 8.75481 & 83.3069 \pm 10.9394 \\
    & CIFAR100 & 20.3631 \pm 7.25594 & 47.1912 \pm 9.08947 & 36.0261 \pm 8.77176 & 56.3531 \pm 14.9311 \\
    & CUB & 0.971828 \pm 0.300213 & 4.24258 \pm 4.07371 & 24.5134 \pm 10.9077 & 4.35464 \pm 1.27898 \\
    & ImageNet & 3.25538 \pm 2.1025 & 15.9979 \pm 9.78166 & 5.643 \pm 4.34291 & 62.7186 \pm 9.49182 \\
    \bottomrule
  \end{tabular}
  \caption{Faithfulness metrics by encoder and dataset, averaged over models.}\label{fig:faithfulness_by_dataset_by_encoder}
\end{table*}

\subsection{Hypothesis Tests} 
For all hypothesis tests we consider as significance level a p-value of $0.05$.
\paragraph{Correlations.}
The p-values of the Pearson Correlation tests\footnote{Pearson correlation tests uses the scipy pearsonr implementation: \url{https://docs.scipy.org/doc/scipy/reference/generated/scipy.stats.pearsonr} } proving correlations between zero-shot accuracy and other metrics can be found in \cref{tab:hypothesis-tests-pearsoncorr-zero-shot}. See the scipy pearsonr implementation\footnote{\url{https://docs.scipy.org/doc/scipy/reference/generated/scipy.stats.pearsonr.html}} for details.

\begin{table*}
  \centering
  \sisetup{table-format=1.4}
  \begin{subtable}{\linewidth}
    \centering
    \begin{tabular}{lS[table-auto-round]rlSr}
      \toprule
      & {Pearson Corr. Coeff.} & {direction} & {significant} & {p-value} & \# {experiments} \\\midrule
      Onto. Sim. 1 (Yago) & 0.389263 & 1 & \checkmark & 0.000001 & 156 \\
      Onto. Sim. 1 (SUMO) & -0.462333 & -1 & \checkmark & 0.000000 & 156 \\
      Onto. Sim. 1 (OpenCyc) & -0.320304 & -1 & \checkmark & 0.000046 & 156 \\
      Hier. Consist. (Yago) & -0.298722 & -1 & \checkmark & 0.000152 & 156 \\
      Hier. Consist. (SUMO) & -0.456285 & -1 & \checkmark & 0.000000 & 156 \\
      Hier. Consist. (OpenCyc) & -0.219829 & -1 & \checkmark & 0.005826 & 156 \\
      tree & 0.732178 & 1 & \checkmark & 0.000000 & 156 \\
      tree soft & 0.767477 & 1 & \checkmark & 0.000000 & 156 \\
      faithfulness & 0.575458 & 1 & \checkmark & 0.000000 & 156 \\
      \bottomrule
    \end{tabular}
    \caption{Correlation with top-1 zero-shot accuracy.}
  \end{subtable}
  \\[1.5\baselineskip]\strut

  \begin{subtable}{\linewidth}
    \centering
    \begin{tabular}{lS[table-auto-round]rlSr}
      \toprule
      & {Pearson Corr. Coeff.} & {direction} & {significant} & {p-value} & \# {experiments} \\\midrule
      Onto. Sim. 1 (Yago) & 0.506095 & 1 & \checkmark & 0.000000 & 156 \\
      Onto. Sim. 1 (SUMO) & -0.542660 & -1 & \checkmark & 0.000000 & 156 \\
      Onto. Sim. 1 (OpenCyc) & -0.404119 & -1 & \checkmark & 0.000000 & 156 \\
      Hier. Consist. (Yago) & -0.611837 & -1 & \checkmark & 0.000000 & 156 \\
      Hier. Consist. (SUMO) & -0.629886 & -1 & \checkmark & 0.000000 & 156 \\
      Hier. Consist. (OpenCyc) & -0.181730 & -1 & \checkmark & 0.023178 & 156 \\
      tree & 0.959624 & 1 & \checkmark & 0.000000 & 156 \\
      tree soft & 0.926583 & 1 & \checkmark & 0.000000 & 156 \\
      \bottomrule
    \end{tabular}
    \caption{Correlation with faithfulness.}
  \end{subtable}
  \caption{\textbf{Hypothesis tests: Zero-shot correlated.}
    Results of the two-sided Pearson correlation test for correlation of top-1 zero-shot accuracy (top) respectively faithfulness (bottom) with other metrics.
    Experiments vary over 13 models, 4 datasets, and 3 leaf embedding methods, using LLM-only text embedding strategy for the non-WordNet-aligned CUB dataset, else the WordNet-only strategy.
  The results also imply the respective one-sided case.}
  \label{tab:hypothesis-tests-pearsoncorr-zero-shot}
\end{table*}

For the per-model results see \cref{tab:hypothesis-tests-tree-inf-zero-shot}.

\begin{table*}

  \centering
  \begin{tabular}{lSlSc}
    \toprule
    score\_b & \multicolumn{2}{r}{tree inference acc.} & \multicolumn{2}{r}{tree inference soft acc.} \\
    & {p-value} & {significant} & {p-value} & {significant} \\
    model &  &  &  &  \\
    \midrule
    ALIGN & 0.000001 & \checkmark & 0.004130 & \checkmark \\
    C RN101 & 0.000001 & \checkmark & 0.669648 & $\times$ \\
    C RN50 & 0.000001 & \checkmark & 0.080912 & $\times$ \\
    C RN50x16 & 0.000001 & \checkmark & 0.021844 & \checkmark \\
    C RN50x4 & 0.000000 & \checkmark & 0.095238 & $\times$ \\
    C RN50x64 & 0.000002 & \checkmark & 0.004662 & \checkmark \\
    C V-B16 & 0.000002 & \checkmark & 0.001635 & \checkmark \\
    C V-B32 & 0.000000 & \checkmark & 0.001999 & \checkmark \\
    C V-L14 & 0.000000 & \checkmark & 0.000020 & \checkmark \\
    C V-L14@336 & 0.000002 & \checkmark & 0.000129 & \checkmark \\
    FLAVA & 0.000000 & \checkmark & 0.000786 & \checkmark \\
    SigLIP 224 & 0.000001 & \checkmark & 0.000345 & \checkmark \\
    SigLIP 384 & 0.000002 & \checkmark & 0.000077 & \checkmark \\
    \bottomrule
  \end{tabular}
  \caption{\textbf{Hypothesis test: Zero-shot $>$ tree inference accuracy.} Results of the one-sided t-tests testing against the 0-hypothesis that top-1 zero-shot accuracy is not different from tree inference / soft tree inference accuracy, instead of being greater. Each test includes more than 10 experiments (different datasets, models, and leaf embedding strategies).}
  \label{tab:hypothesis-tests-tree-inf-zero-shot}
\end{table*}

\paragraph{Zero-shot vs. Tree inference}
The left-sided t-test against
\begin{itemize}
  \item 0-hypothesis: the difference between zero-shot accuracy and tree inference is normally distributed around 0, and
  \item 1-hypothesis: zero-shot is greater, i.e., the difference is $<0$,
\end{itemize}
on 273 experiments yielded a mean difference of $27.51$, corrected standard deviation of $18.95$, and a standard error of $1.15$, resulting in a
p-value of $\num{2.23e-69}$ (strongly significant).

The left-sided t-test against
\begin{itemize}
  \item 0-hypothesis: the difference between zero-shot accuracy and \emph{soft} tree inference accuracy is normally distributed around $0$, and
  \item 1-hypothesis: zero-shot is greater, i.e., the difference is $<0$
\end{itemize}
on 273 experiments yielded a mean difference of $10.89$, corrected standard deviation of $17.27$, and a standard error of $1.0455$, resulting in a
p-value of $\num{6.52e-22}$ (strongly significant).

\subsection{Comparison of Distance to LCA by Encoder}
\Cref{tab:appendix-dist-lca} shows more detailed results on the distance of a tree traversal prediction from the LCA of the leaf prediction with the ground truth, comparing different encoders and datasets.

\begin{table*}
  \centering
  \includegraphics[width=.3\linewidth]{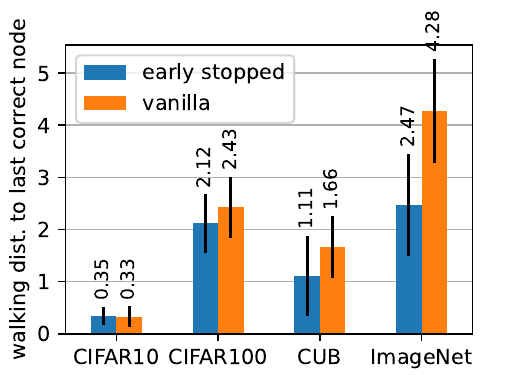}
  \sisetup{table-format=3.3(2)}
  \begin{tabular}{llS|S|S|S}
    \toprule

    {encoder} & {method} & {CIFAR10} & {CIFAR100} & {CUB} & {ImageNet} \\\midrule[\heavyrulewidth]
    \multirow[c]{2}{*}{image} & early stop & 0.0694961 \pm 0.0338999 & 0.189842 \pm 0.0283582 & 0.1702 \pm 0.027191 & 0.178396 \pm 0.0151002 \\
    & full & 0.0602428 \pm 0.0349731 & 0.197746 \pm 0.028518 & 0.198744 \pm 0.0246555 & 0.226635 \pm 0.0248478 \\\midrule
    \multirow[c]{2}{*}{text} & early stop & 0.0857133 \pm 0.0408511 & 0.130624 \pm 0.039292 & 0.0138058 \pm 0.00810018 & 0.034903 \pm 0.0226702 \\
    & full & 0.0895872 \pm 0.049406 & 0.176245 \pm 0.0443551 & 0.0490746 \pm 0.00492161 & 0.104375 \pm 0.0361168 \\\midrule
    \multirow[c]{2}{*}{text \& image} & early stop & 0.0714849 \pm 0.0357628 & 0.231245 \pm 0.0558676 & 0.0704347 \pm 0.0200695 & 0.1468 \pm 0.0132536 \\
    & full & 0.063654 \pm 0.0372978 & 0.253186 \pm 0.0642276 & 0.109412 \pm 0.0162623 & 0.241841 \pm 0.0313628 \\
    \bottomrule
  \end{tabular}
  \caption{Detailed normalized results on how much early stopping improves upon predicting correct (super-)categories: normalized tree-distance scores ($\downarrow$) between the predicted node and the last correct node (LCA of the ground truth with the leaf predicted by tree traversal); averaged over models and grouped by encoder.}\label{tab:appendix-dist-lca}
\end{table*}

\subsection{Comparison of Text Embedding Strategies}  
To embed concepts, we newly used definition and synonym information from WordNet to create an ensemble of textual descriptions (details in \cref{sec:appendix-concept-bank}). This splits our textual description list into the part created solely by means of WordNet, and the part using an LLM (with WordNet disambiguation hints) to add further descriptions.
To check whether the LLM-based textual descriptions from \shortcite{ma2025does} are needed on top of the WordNet ones, we compared using the WordNet-only (solely the term, its synonyms and definition), LLM-only (solely descriptions of visual features), and combined parts of the list.

\begin{figure*}
  \centering
  \includegraphics[width=.3\linewidth]{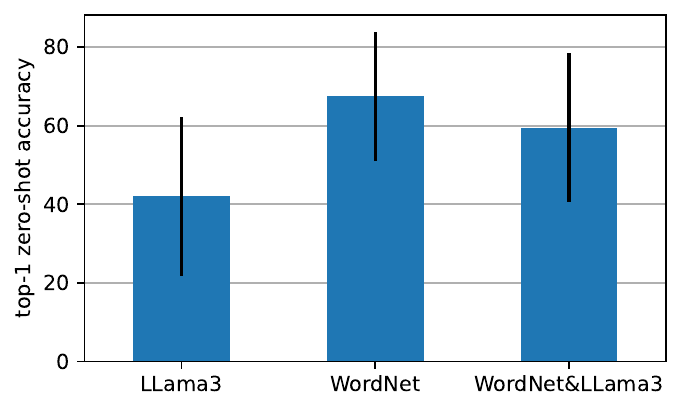}
  \includegraphics[width=.3\linewidth]{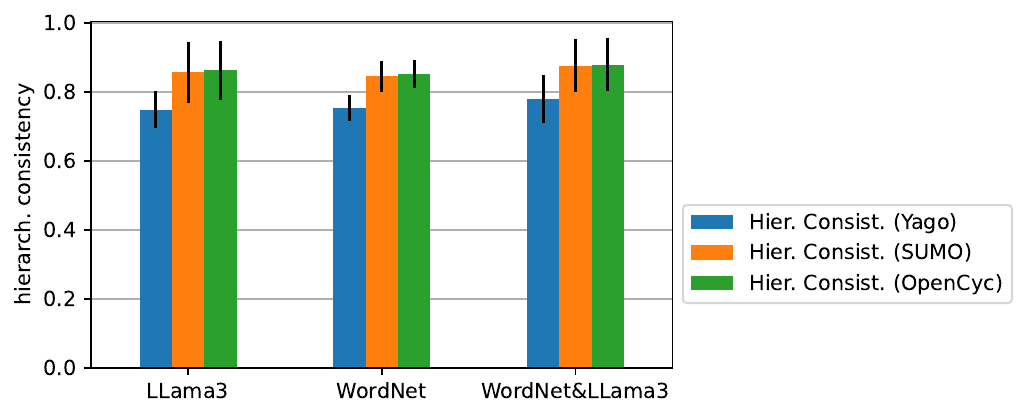}
  \\\includegraphics[width=\linewidth]{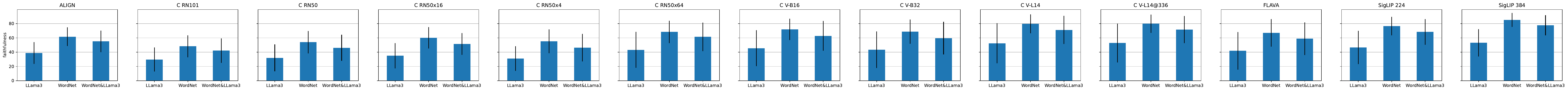}
  \caption{\textbf{Comparing text embedding strategies:}
    Top 1 zero-shot accuracy per text embedding strategy
    on CIFAR-10, CIFAR-100, and ImageNet and for different leaf embedding strategies (averaged).
    Top: averaged over models, bottom: per model.
  }
  \label{fig:zero-shot-augmenters}
\end{figure*}

\paragraph{Results.}
The detailed results can be found in \cref{fig:zero-shot-augmenters}. We tested on text-embedded leaf concepts discriminativity of the embedded concepts (via zero-shot performance), and plausibility of resulting trees (via directed SUMO scores).
In summary, the LLM-only version is inferior with respect to the versions using WordNet directly, and even significantly so for zero-shot performance.

\subsection{Comparison of Datasets}
The detailed results showing the negative influence of the number of leaf nodes on faithfulness are shown in \cref{fig:faithfulness-by-dataset}.
\cref{fig:faithfulness_by_dataset_by_encoder} shows this more finegrained showing faithfulness metrics per dataset per encoder. It becomes clear how text embeddings are more impacted by the faithfulness drop in larger trees. This  indicates less stable semantic structures.

\begin{figure*}
  \centering
  \includegraphics[width=\linewidth]{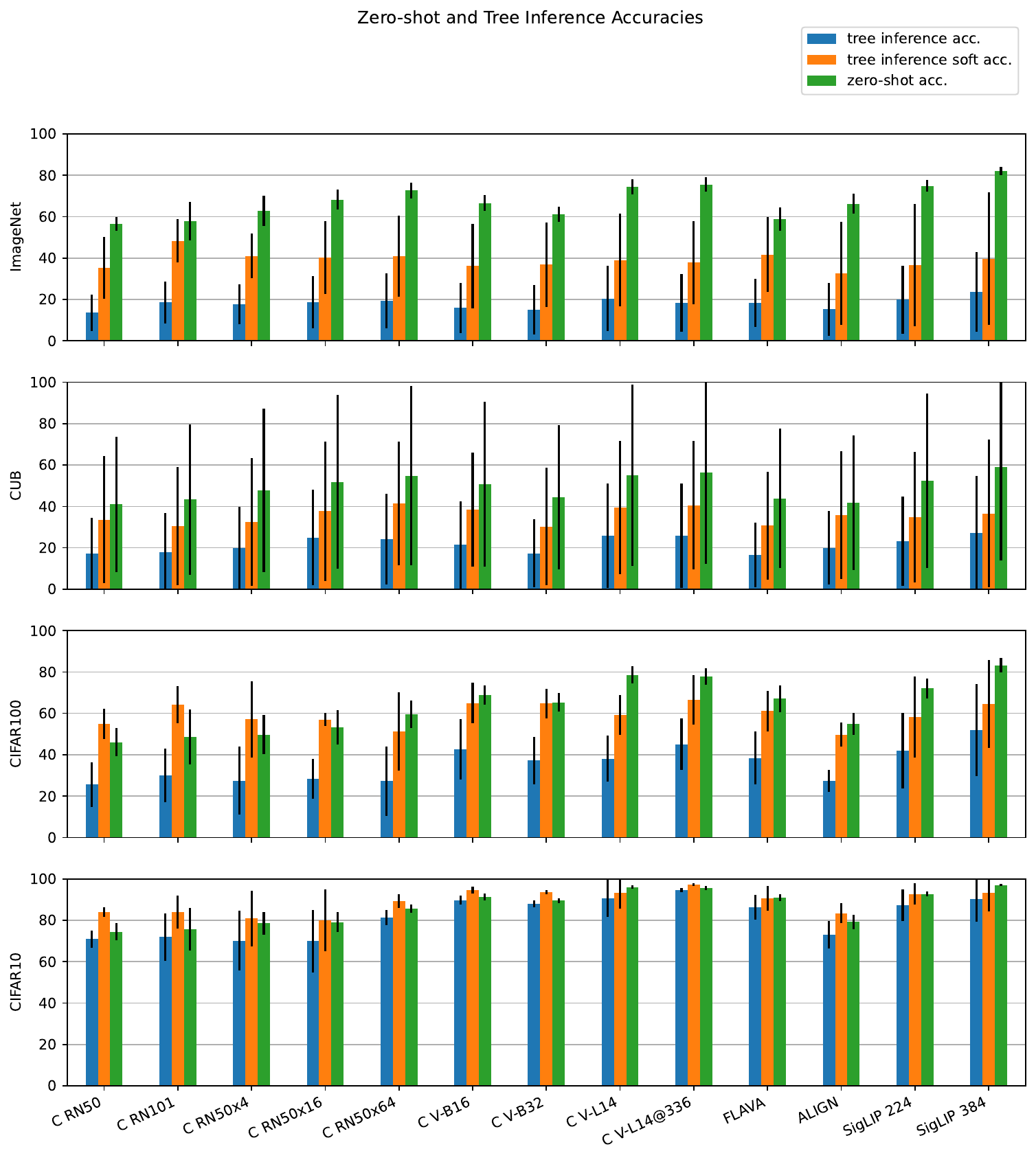}
  \caption{\textbf{Comparison of datasets:} Faithfulness only per dataset and model averaged over text embedding strategies (WordNet, WordNet+LLM).
  }
  \label{fig:faithfulness-by-model-by-dataset}
\end{figure*}

The effect differs between the different leaf embedding strategies, but shows the same trend.
\Cref{fig:faithfulness-by-model-by-dataset} concentrates on comparing faithfulness per model across datasets.
While the more complex datasets (CUB, ImageNet) show few differences between the faithfulness of different models, slight differences can be observed for the smaller CIFAR10\&100.

\subsection{Comparison of Encoders}
\begin{figure*}
  \centering
  \includegraphics[width=.5\linewidth]{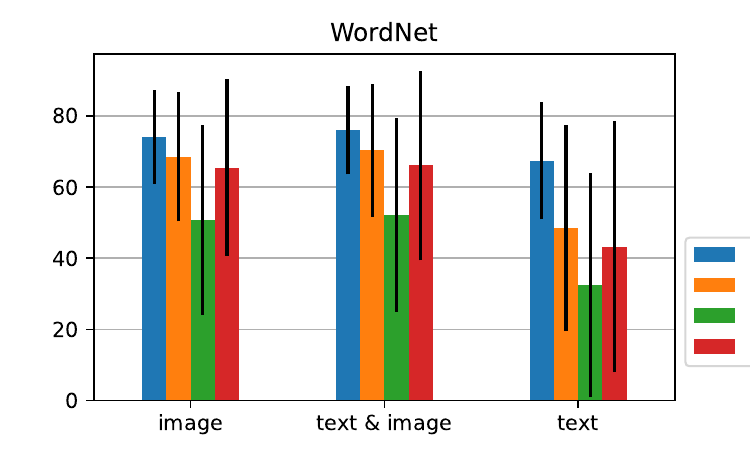}
  \caption{\textbf{Comparing leaf embedding strategies:} Faithfulness metrics averaged across datasets and models.}
  \label{fig:faithfulness-by-leaf-mode-by-augmenter}
\end{figure*}

In \cref{fig:faithfulness-by-leaf-mode-by-augmenter} the impact of different text embedding strategies is shown on the overall faithfulness, when varying the leaf embedding strategy and the text embedding strategy (WordNet, WordNet+LLM).

\subsection{Comparison of Metrics: Soft Tree-traversal Accuracy versus Tree-traversal Accuracy}

\begin{table}
  \centering
  \begin{tabular}{lSS}
    \toprule
    & {\# classes} & {average maximum tree depth} \\
    \midrule
    CIFAR10 & 10 & 6.30769 \pm 0.995347 \\
    CIFAR100 & 100 & 20.7094 \pm 3.68141 \\
    CUB & 200 & 25.0128 \pm 5.84984 \\
    ImageNet & 1000 & 51.2292 \pm 22.1267 \\
    \bottomrule
  \end{tabular}
  \caption{Average maximum depth of trees extracted for different leaf class sets.
  Averaged over models and encoder modality.}\label{tab:average-max-tree-depth}
\end{table}

The main paper results showed partly strong drops in top-1 tree inference accuracy.
A common way to further evaluate zero-shot performance is to consider top-5 or higher performance, i.e., use the induced ranking of leaf nodes to assess whether the correct one is under the top-5 ranked one.

While the tree inference also produces rankable confidence scores at each comparison in the traversal, obtaining a ranking directly from the comparisons would require a large number of hyperparameters (the confidence scales in each tree level).
We therefore define \textbf{soft tree inference accuracy}:
The score is the percentage of the path from root to the correct leaf that the tree's prediction traversed. This is $0$ in case a wrong turn was taken right at the root node, $1$ for an exact match, and has a high value if the predicted child was a sibling.
For example, for a balanced binary hierarchy of CIFAR10 leaf classes and a root-to-leaf path of length 3, predicting a direct sibling would result in a score of $0.67$ instead of $0$ for that test sample. The true average tree depths are shown in \cref{tab:average-max-tree-depth}.
The final soft tree inference accuracy is the mean of the soft accuracies per test sample.

The hypothesis is that the tree predictions are not top-1 accurate, but still close to the ground truth (the \enquote{wrong turn} is taken close to the leaves).
This would imply a high soft tree inference accuracy that is closer to the top-1 zero-shot performance than to the tree inference accuracy.

\begin{figure*}
  \centering
  \includegraphics[width=\linewidth]{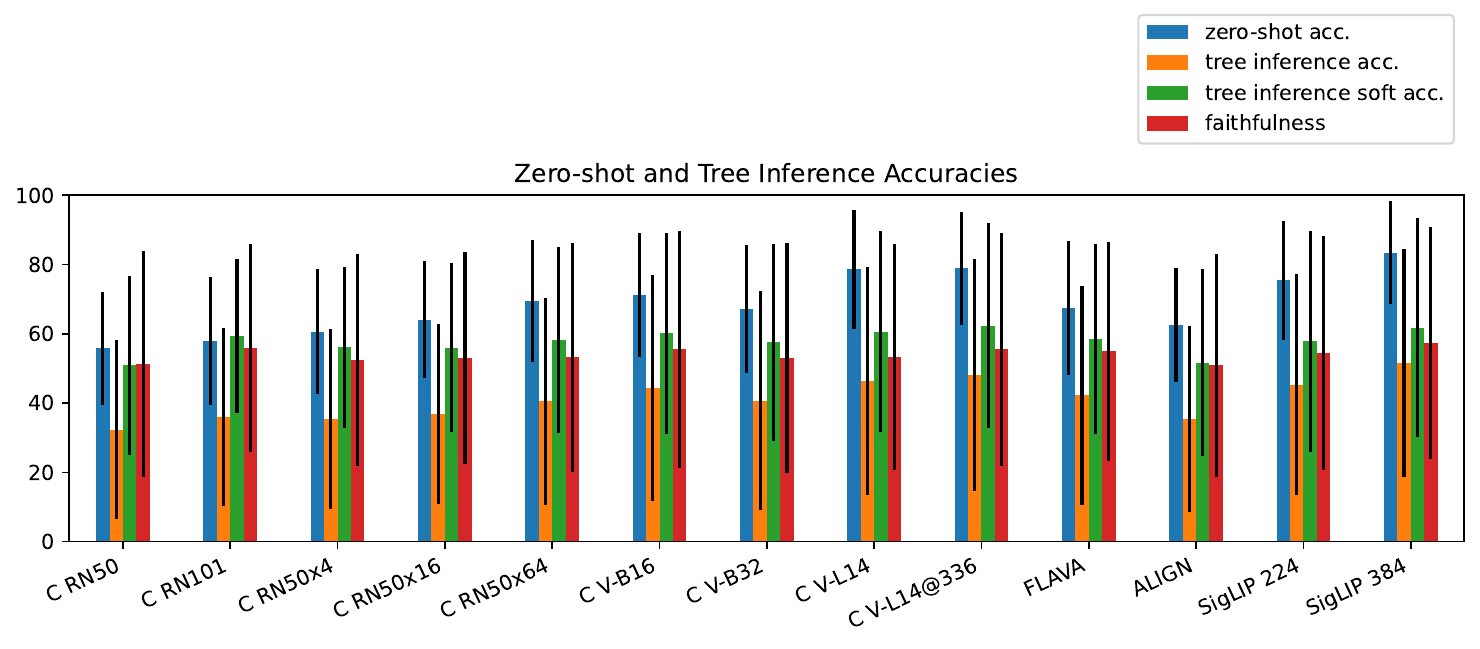}
  \includegraphics[width=\linewidth]{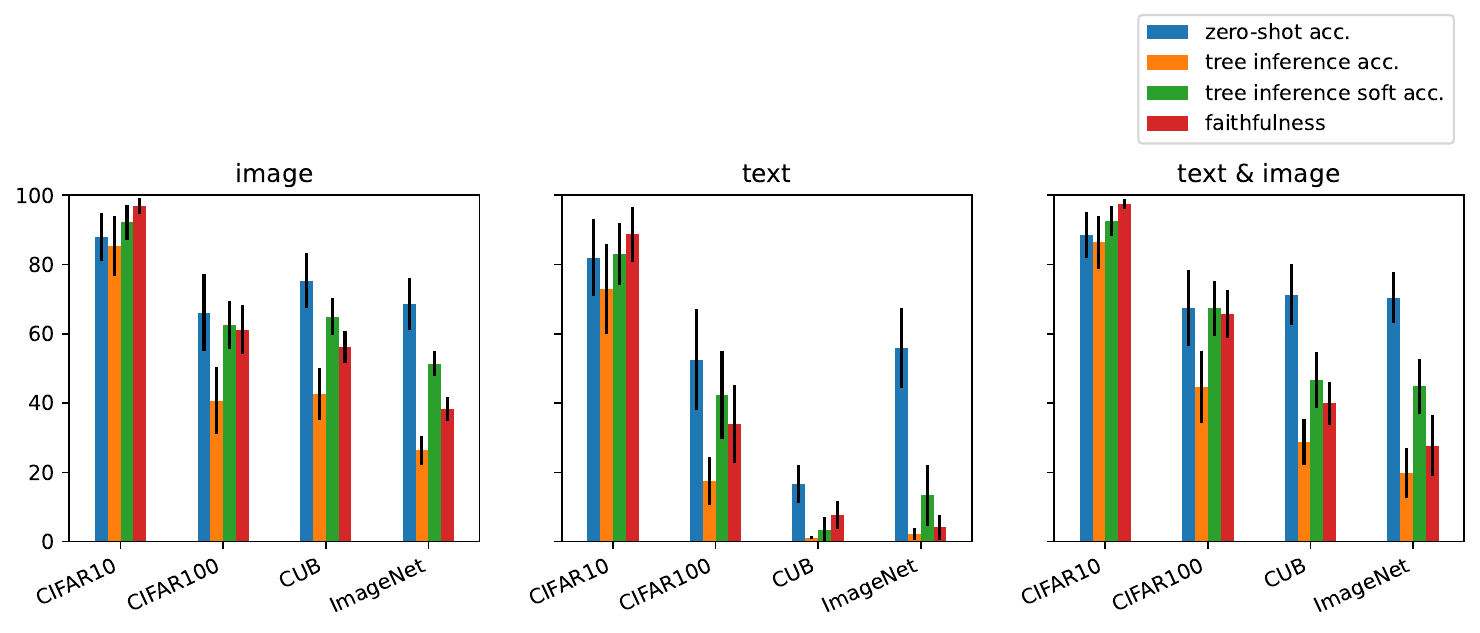}
  \caption{\textbf{Explaining the faithfulness gap:}
    Comparison of faithfulness metrics, in particular soft tree inference accuracy, tree inference accuracy, and zero-shot accuracy, averaged over models, datasets, text embedding strategies WordNet \& WordNet+LLM, and leaf embeddings trategies.
    Top: by model; bottom: by dataset and leaf embedding strategy.
  }
  \label{fig:faithfulness-soft}
\end{figure*}

\paragraph{Results.}
\Cref{fig:faithfulness-soft} shows soft tree inference results by model, and by dataset and leaf embedding mode.
It is observed that soft tree accuracy explains away a substantial part of the gap between zero-shot accuracy and tree inference accuracy (a \enquote{wrong turn} was taken way down the tree). However, for the datasets with larger amounts of leaves, a substantial gap remains, indicating that \textbf{still the tree inference lacks faithfulness to the zero-shot comparisons}, invalidating the hypothesis.

\section{More Details on Experimental Settings}

\subsection{Used Implementations}
\paragraph{Hierarchical Clustering.}
For hierarchical clustering we use standard agglomerative clustering \cite{wardjr.1963hierarchical}, with the \texttt{scikit-learn}\footnote{\tiny\url{https://scikit-learn.org/stable/modules/generated/sklearn.cluster.AgglomerativeClustering}}
(version 1.6.1) implementation.
Following CLIP training, the tree extraction uses \textbf{cosine similarity} both for the affinity in hierarchical clustering as well as for tree inference (see below), and the respective standard setting of \textbf{average} linkage.

\paragraph{Models.}
CLIP: \url{https://github.com/openai/CLIP}
using backbones
ViT-B/16, ViT-L/14@336px, RN50x4, RN50x16, 
ViT-B/32, ViT-L/14, RN50, RN101, RN50x64.  
\\ALIGN: \url{https://huggingface.co/kakaobrain/align-base}
\\FLAVA: \url{https://huggingface.co/facebook/flava-full}
\\SigLIP: \url{https://huggingface.co/google/siglip-base-patch16-{bb}}
for backbones\,(bb) 224, 384.

\paragraph{Tree Edit Distance.}
For the normalized UTED we use the \texttt{edist} implementation from \cite{Paassen2015EDM}.

\subsection{Hardware Settings and Runtime}
Experiments were run on an NVIDIA H100 GPU (95GB VRAM) and an AMD EPYC 9654 96-Core Processor with 1.5TB memory.
For the ontology modification experiments, we used an NVIDIA A100 GPU(80GB VRAM) and an four Intel(R) Xeon(R) Gold 6242 CPU @ 2.80GHz with 1.5TB memory.
Calculation of the validation metrics scales with the size of the ontology graph. For SUMO, calculations took approximately 1, 2, and 20 seconds for CIFAR10, CIFAR100, and ImageNet respectively. For OpenCyc these times increased to 4, 8, and 80 seconds. Finally, for YAGO, the calculations required approximately 15, 30, and 300 seconds respectively.





\subsection{Ontology Verification Settings: Graph Construction \& Metrics}

\phantomsection
\subsubsection{Graph Construction and Relations}

To evaluate the semantic validity of the VLM-derived hierarchies, we leverage three distinct expert knowledge bases: \textbf{SUMO}, \textbf{OpenCyc}, and \textbf{YAGO 3}. Since these ontologies differ in structure and density, we first normalize them into a consistent Directed Acyclic Graph (DAG) format and then apply a unified scoring metric (Mode 1).

For each ontological framework, we construct a taxonomic graph $\mathcal{G} = (V, E)$ where an edge $u \to v$ represents a direct hypernymy relation (``$u$ is a parent of $v$''). The specific predicates used to define this connection differ by ontology:
\begin{itemize}
  \item \textbf{SUMO:} We utilize the explicit \Concept{Subclass} and \Concept{InstanceOf} predicates defined in the KIF source files.
  \item \textbf{OpenCyc:} We extract the taxonomic backbone using the Semantic Web standard predicates \Concept{rdfs:subClassOf} and \Concept{rdf:type} from the OWL export.
  \item \textbf{YAGO 3:} We filter the massive YAGO knowledge base to isolate the English taxonomic backbone. We utilize \Concept{rdfs:subClassOf} and \Concept{rdf:type} relations, strictly excluding time-dependent, spatial, or meta-logical relations to ensure a clean semantic hierarchy.
\end{itemize}

\paragraph{Cycle Breaking and DAG Enforcement.}
While expert ontologies like SUMO are largely hierarchical, large-scale knowledge bases (especially those derived from Wikipedia like YAGO) often contain logical cycles (e.g., $A \to B \to C \to A$), which render hierarchical distance metrics undefined.
To enable valid Lowest Common Ancestor (LCA) calculations, we enforce a Directed Acyclic Graph (DAG) structure for all three ontologies. We employ a depth-first cycle detection algorithm; upon detecting a cycle, we iteratively remove the "back-edge" (the edge returning to an already visited ancestor) until the graph is strictly hierarchical. This preprocessing ensures that every node in $\mathcal{G}$ has a finite path to the root entity.

\subsubsection{Metrics}
\paragraph{Alternative Consistency Score: Cluster Consistency.}
To quantify how well a decision tree node $\Concept{P}$ represents a meaningful semantic cluster, we identify the Lowest Common Ancestor (LCA) of its mapped children in the ontology, denoted as $\Concept{L}_{\Concept{P}}$.
Unlike the undirected score used in previous iterations, we employ a \textbf{robust inverse-decay metric} that accounts for the varying semantic density of the different ontologies.
The validity score for a parent $\Concept{P}$ is calculated as:
\begin{gather}
  \text{S1}(\Concept{P}) =
  \begin{cases}
    1.0 & \text{if } \Concept{P} \equiv \Concept{L}_{\Concept{P}} \\
    \frac{1}{1 + k \cdot d_{\mathcal{T}}(\Concept{P}, \Concept{L}_{\Concept{P}})} & \text{if path exists} \\
    0 & \text{otherwise}
  \end{cases}
\end{gather}
where:
\begin{itemize}
  \item $d_{\mathcal{T}}(\Concept{P}, \Concept{L}_{\Concept{P}})$ is the shortest path distance between the tree parent and the ontological LCA.
  \item $k$ is a \textit{density scaling factor} derived from the effective semantic depth of the ontology. Based on the empirical average leaf depth across our target ontologies ($D_{eff} \approx 4.5$), we set $k \approx 0.225$. This ensures the score decays to $0.5$ at the "semantic horizon" of approximately 4-5 hops.
\end{itemize}

This metric penalizes clusters where the purported parent is semantically distant from the true common ancestor of its children, while being robust to the different granularities of SUMO, OpenCyc, and YAGO.

\begin{figure}[ht]
  \centering
  \begin{subfigure}{0.48\linewidth}
    \centering
    \includegraphics[width=\linewidth]{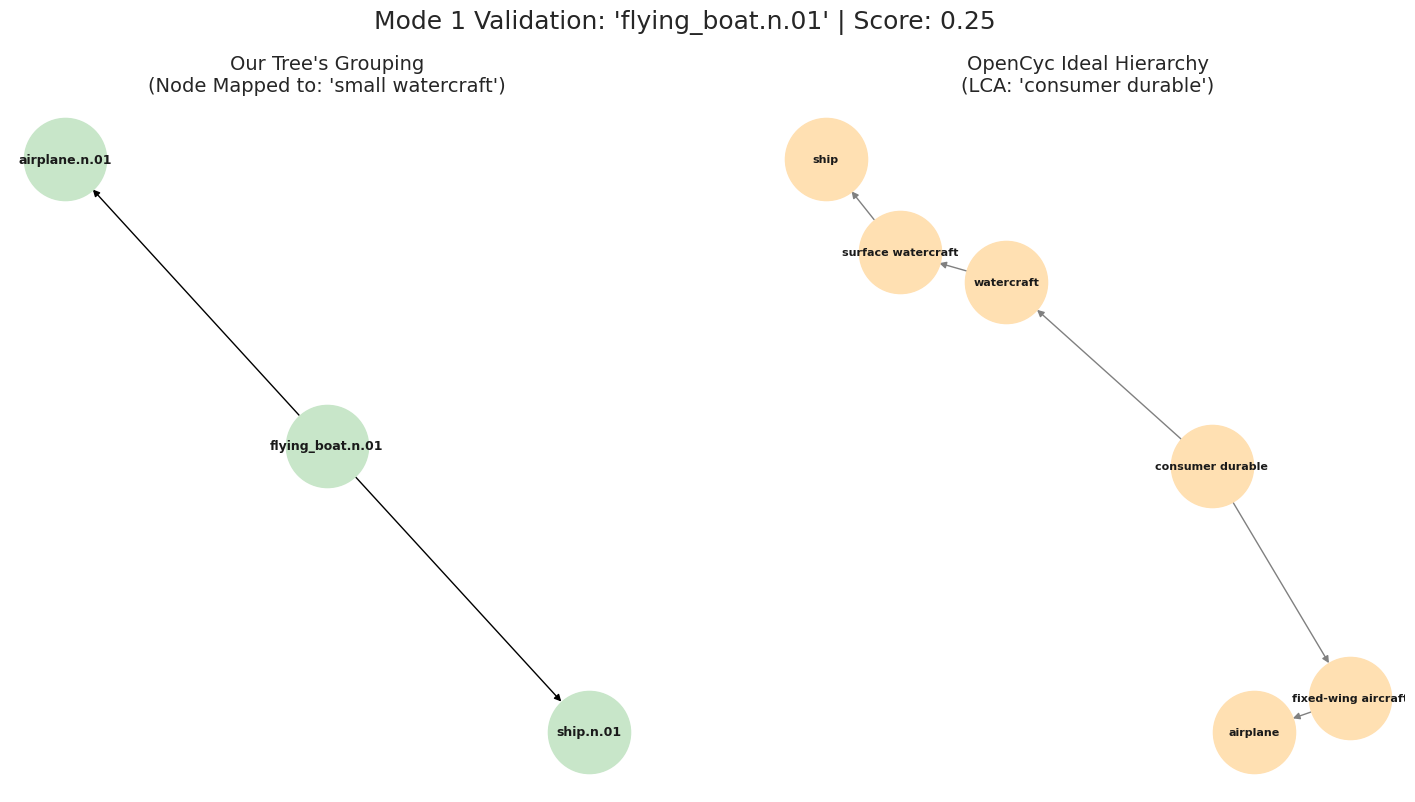}
    \caption{\textbf{Mode 1 (Cluster Consistency).} Evaluates if a decision node forms a valid semantic group based on the LCA of its children.}
    \label{fig:sumo-dists1}
  \end{subfigure}
  \hfill 
  \begin{subfigure}{0.48\linewidth}
    \centering
    \includegraphics[width=\linewidth]{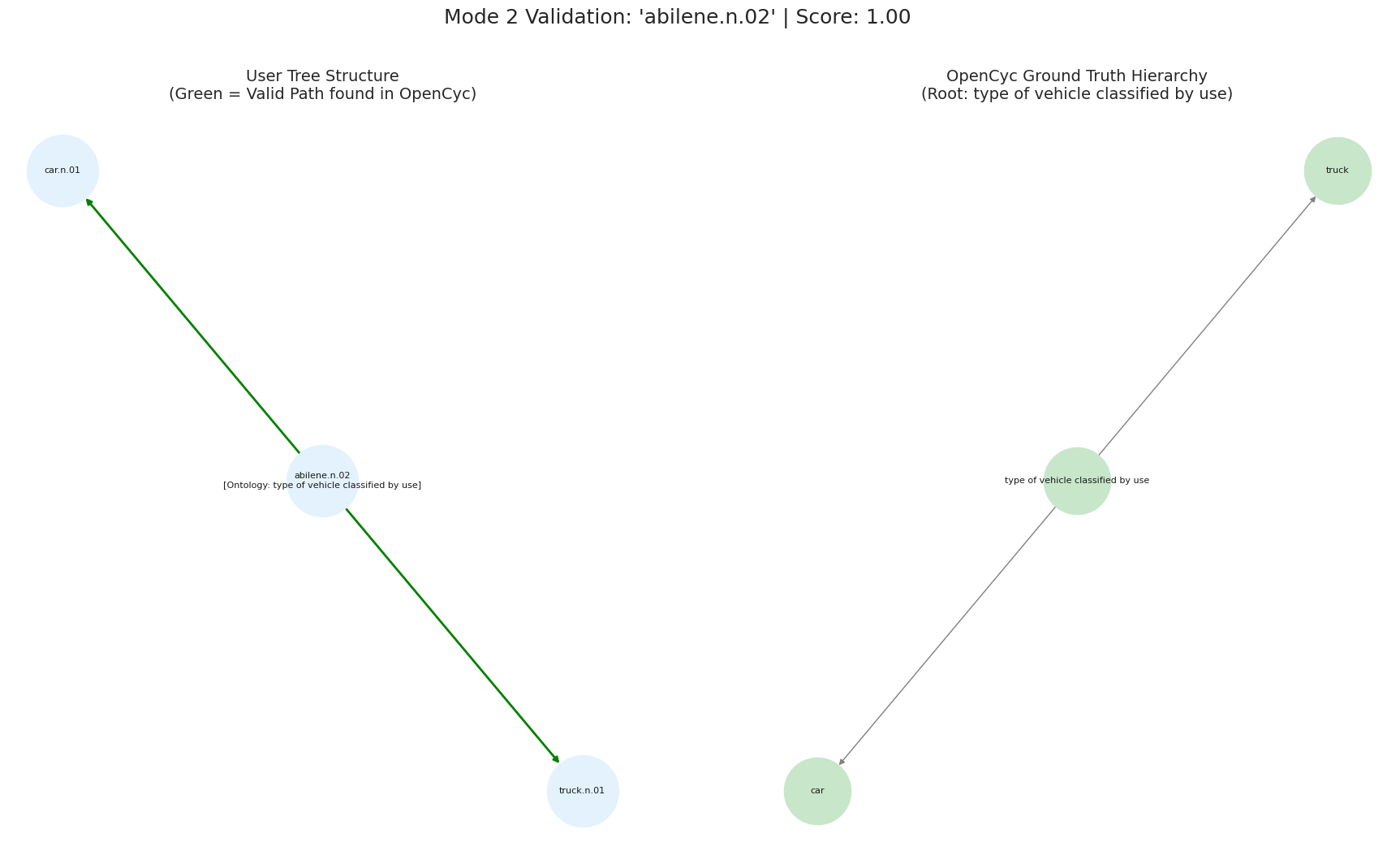}
    \caption{\textbf{Mode 2 (Hierarchical Consistency).} Validates structural integrity by verifying directed paths between inferred parent and child concepts.}
    \label{fig:sumo-dists2}
  \end{subfigure}

  \caption{Illustration of the steps involved to calculate the ontology validation metrics.}
  \label{fig:sumo-dists}
\end{figure}

An illustration of the steps involved to calculate each of the SUMO validation metrics is shown in \cref{fig:sumo-dists}.

\begin{figure*}
  \centering
  \hfill
  \includegraphics[width=.45\linewidth]{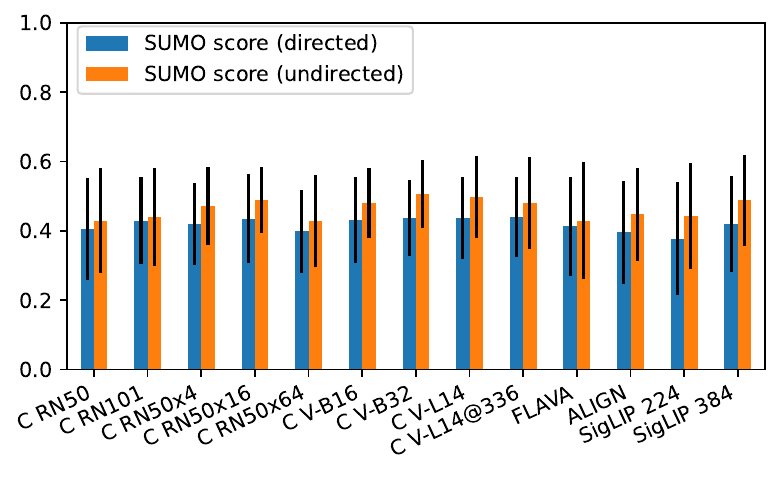}\hfill
  \includegraphics[width=.45\linewidth]{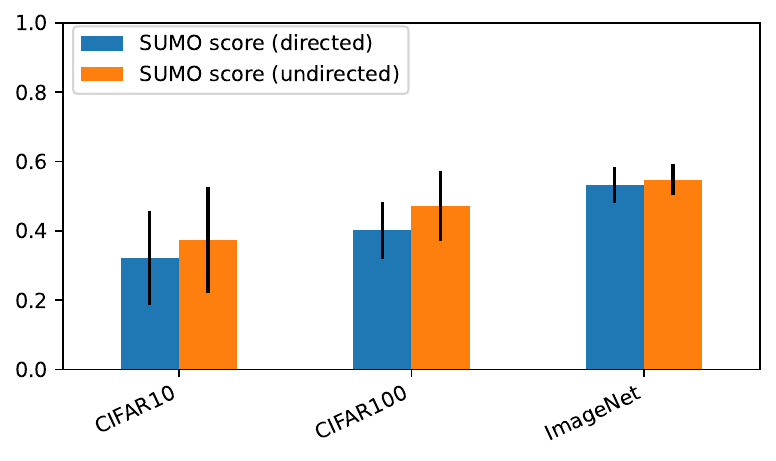}\hfill
  \strut\\\includegraphics[width=.6\linewidth]{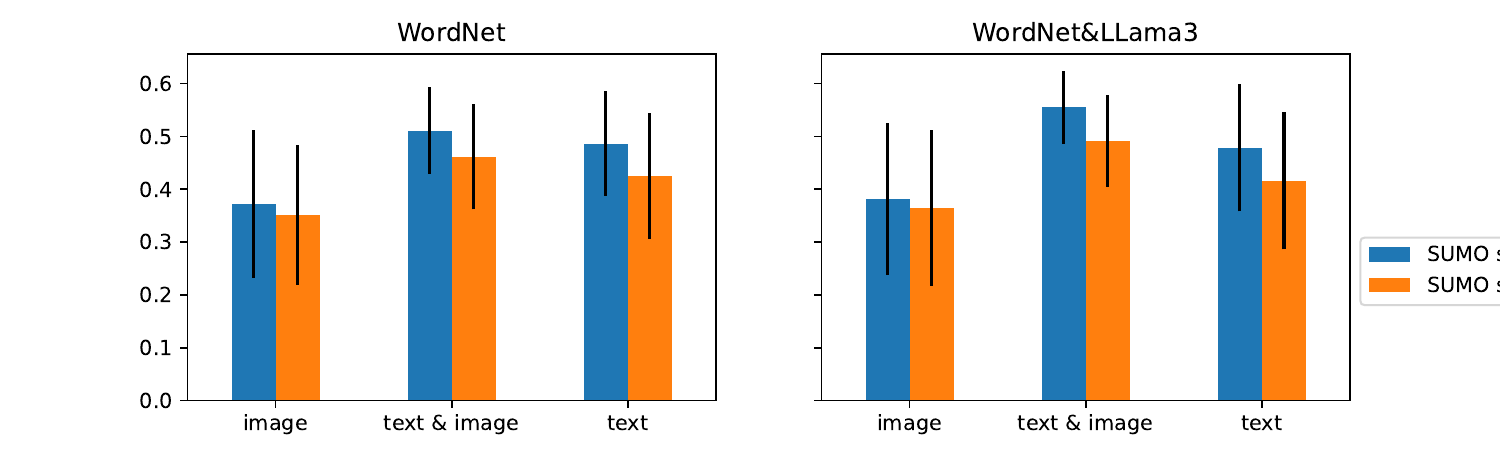}
  \\\includegraphics[width=\linewidth]{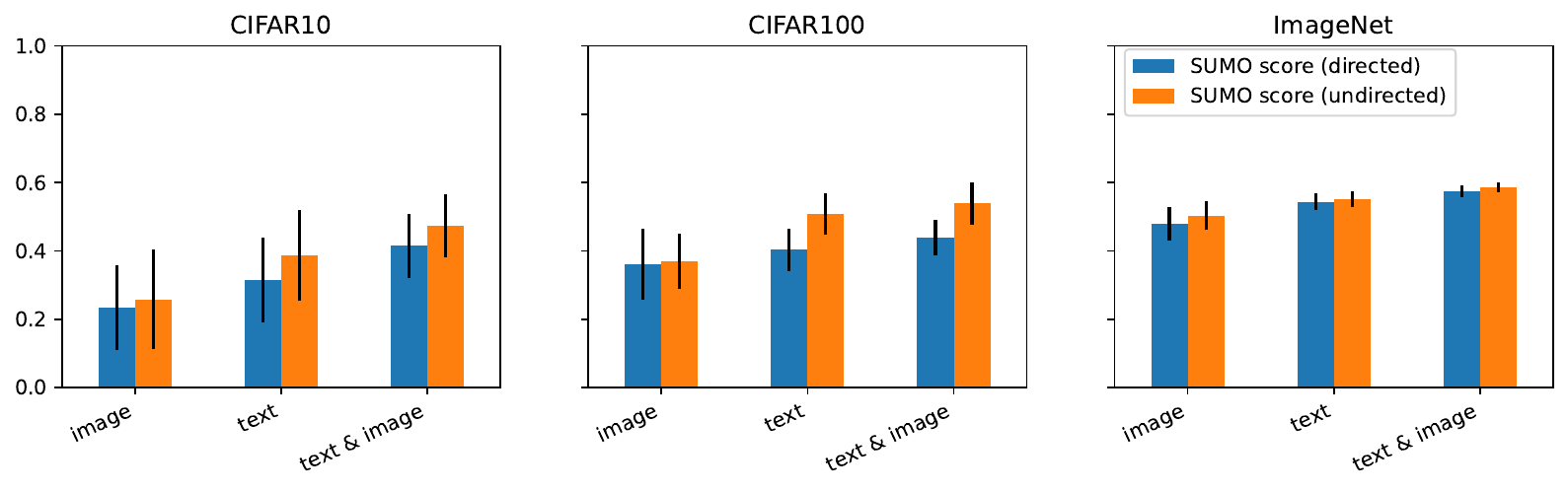}
  \\\includegraphics[width=\linewidth]{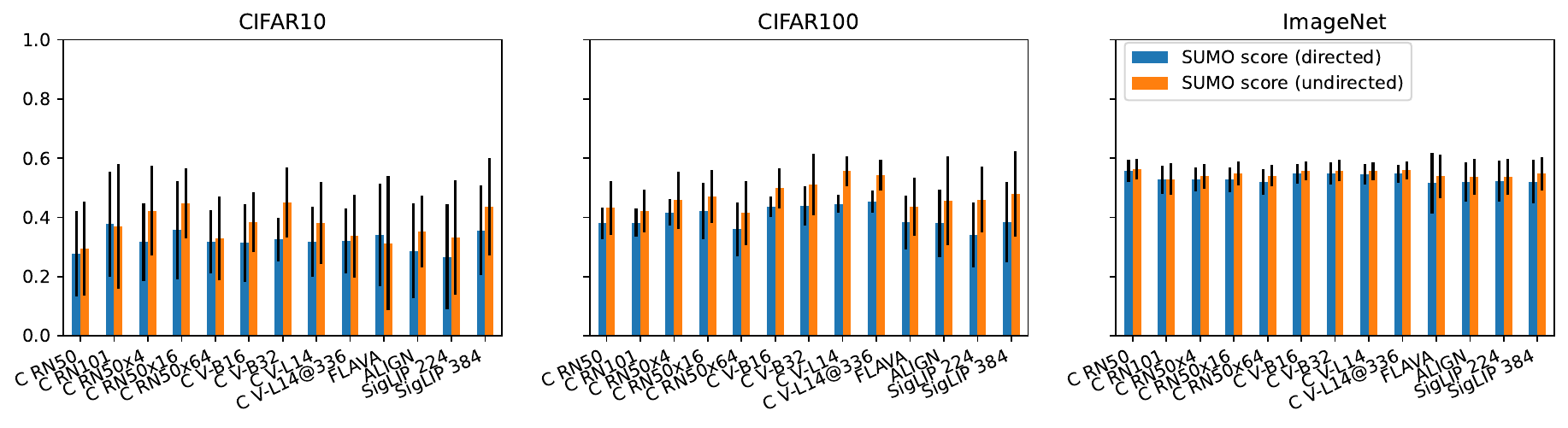}
  \caption{Comparison of the two SUMO validity scores.
  If not stated otherwise, averaged over models, datasets, leaf and text embedding strategies.}
  \label{fig:sumo-validity-score-comparison}
\end{figure*}

\paragraph{Comparison of the local scores.}
A comparison of undirected and directed score is shown in \cref{fig:sumo-validity-score-comparison}.
Concretely, we test the two scores for the SUMO ontology and the trees resulting from 234 experiments (3 standard datasets, 13 models, 2 text and 3 leaf embedding strategies).
This shows that the two are significantly correlated with a Pearson correlation coefficient of $0.86$, and p-value lower than $10^{-6}$.
Altogether, the influence of the too specific parent scenarios seems to be neglectable, with the stricter scores being consistently only slightly smaller and showing the same trends.
We therefore only report the less strict validity score.

\begin{figure}
  \centering
  \includegraphics[width=\linewidth]{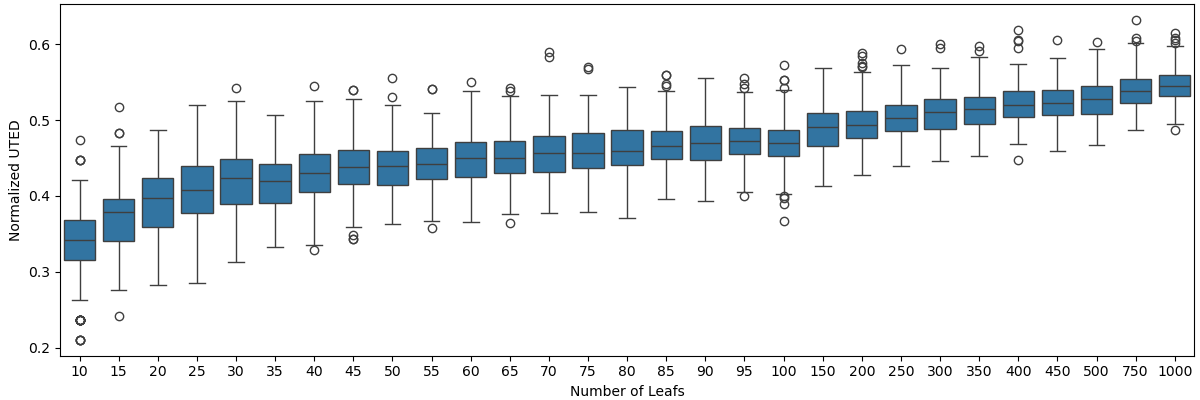}
  \caption{Empirical approximation of the expected normalized UTED across randomly generated binary trees with identical leaf sets. Results based on 200 runs for varying numbers of leaf nodes.
  }
  \label{fig:random_empirical_uted}
\end{figure}

\paragraph{Intuition behind UTED.}
Apart from the validity scores described above, we also use unordered tree edit distance (UTED) to compare trees.
\Cref{fig:random_empirical_uted} collects empirical UTED results when comparing binary trees of different size that each have a random assignment of the node labels to the nodes. Notably, this random setting achieves UTED values of approximately $0.35$ for very few leaves, stabilizing around $0.5$ for mid to large numbers of leaves ($100$ to $1000$, mind the change of scale between $0$-$100$ and $100$-$1000$).

\subsection{Concept Bank Embeddings \& Hyperparameter Study}\label{sec:appendix-concept-bank}  
In this section we describe in more detail how the embeddings for image and, in particular, textual leaf and concept bank concepts were obtained.
Consider a concept $\Concept{C}$ that is specified via image or textual samples $x\in\text{Spec}(\Concept{C})$.
Similar to other probing techniques based on the interpolatability assumption, we assign to it the embedding space vector
\begin{gather}
  e(\Concept{C}) \coloneqq \mean_{x\in\text{Spec}(\Concept{C})}F(x).
\end{gather}
We consider as sources for the specification set $\text{Spec}(\Concept{C})$ either a set of \textbf{images} containing $\Concept{C}$ taken from a classification dataset (for leaves only); or a set of descriptive \textbf{textual prompts}; or the \textbf{mean} of a text-based embedding and an image-based one (also leaves only).

For generating the textual prompts we enhance the recent ensembling-based technique of \shortcite{ma2025does} by ontology information. They obtain $e(\Concept{C})$ as the weighted mean of text encodings of the static prompt \enquote{a photo of \textit{concept}}, and of LLM-generated visual feature descriptions.
It is assumed that all ontology concepts map to a set of textual synonyms and a definition, e.g., in the WordNet lexical database \cite{fellbaum2010wordnet}.
We create static prompts for all of these (\textit{WordNet} part of $\text{Spec}(\Concept{C})$), and also
include them in the LLM query for better disambiguation
(\textit{LLM} part). See supplementary for details.
This differs from \cite{ma2025does} in that they do not attend ambiguity, nor add additional synonyms which required them to use a weighted mean, upvoting the phrase containing the actual term.
Note that this technique combines the prompt enhancement from the original CLIP inference \cite{radford2021learning} with dictionary-enhanced prompting as of \shortcite{ge2023improving}, and with ensembling ideas from \shortcite{roth2023waffling} and the recent feature description based ensembling of \shortcite{ma2025does}.

\subsubsection{LLM Queries}
For generating the textual prompts we enhance the technique by \shortcite{ma2025does}, considering not only an LLM but also the ontology as source of textual descriptions.
The ontology source assumes that the concepts in $\mathcal{C}$ map to a synonym set in the WordNet lexical database \cite{fellbaum2010wordnet}. Textual descriptions are derived from the synonymous terms and definition as
\begin{enumerate}[label=(\arabic*)]
  \item \enquote{a photo of \textit{term}} for all synonymous terms,
  \item \enquote{a photo showing a \textit{term1}, also described as a \textit{term2}, \dots, or a \textit{termN}}, and
  \item \enquote{a photo of a \textit{definition}}\footnote{%
      Removing/adapting the article where needed, and for adjectives using \enquote{a photo of something that is \textit{def}}.
    }.
\end{enumerate}
As a second source we query an LLM (Llama3) to create a list of visual features that apply to the concept, such as \enquote{whiskers}, and \enquote{pointed ears} 
for \Concept{cat}.
To ensure non-ambiguity, the LLM prompt is enhanced as well by the synonyms and definition from WordNet, as detailed below.
$e(\Concept{C})$ is then obtained as the equal weighted mean of all textual descriptions from both sources.
Note that this technique combines the prompt enhancement from the original CLIP inference \cite{radford2021learning}, with dictionary-enhanced prompting as of \shortcite{ge2023improving}, and with ensembling ideas from \shortcite{roth2023waffling} and the recent feature description based ensembling of \shortcite{ma2025does}.

The rest of this section describes further details on the LLM-based query to create a text specification of a concept.
Prompting was done on the Llama3 model \cite{grattafiori2024llama}\footnote{\url{https://ollama.com/library/llama3}}
via the ollama library\footnote{\url{https://github.com/ollama/ollama-python}},
and with temperature of $0$ to ensure reproducibility.
The preceeding system prompt was
\begin{quote}
  \enquote{%
  You are an expert in visual recognition. Describe the visible traits that would help someone recognize an object in a photo.}
\end{quote}
The used prompt for a concept with a specified description term (the synonym set name in WordNet) was:
\begin{quote}
  \enquote{%
    Q: What are useful visual features for distinguishing a lemur in a photo?
    \\A: There are several useful visual features to tell there is a lemur in a photo:
    \\- four-limbed primate
    \\- black, grey, white, brown, or red-brown
    \\- wet and hairless nose with curved nostrils
    \\- long tail
    \\- large eyes
    \\- furry bodies
    \\- clawed hands and feet
    \\~
    \\Q: What are useful visual features for distinguishing a television in a photo?
    \\A: There are several useful visual features to tell there is a television in a photo:
    \\- electronic device
    \\- black or grey
    \\- a large, rectangular screen
    \\- a stand or mount to support the screen
    \\- one or more speakers
    \\- a power cord
    \\- input ports for connecting to other devices
    \\- a remote control
    \\~
    \\Q: What are useful features for distinguishing a \emph{term} in a photo? \emph{definition}
  \\A: Please list several useful visual features to tell there is a \emph{term} in a photo. Format your answer as a bullet point list starting with dash (-).}
\end{quote}
The definition prompt part, if a matching WordNet synonym set is available, is generated as
\begin{quote}
  \enquote{%
    Here a \textit{term},
    also known as \textit{alternative term1}, \textit{alternative term2}, \dots, or \textit{alternative termN}
  is defined as: \textit{def}.}
\end{quote}
where \textit{def} is the WordNet synonyn set definition, and the alternative terms are the synonyms. The prompt is respectively simplified if no synomyms are available.
Articles were changed to 'an' in presence of respective vocals.
For adjectives, \enquote{a \textit{term} in a photo} is replaced by \enquote{whether something in a photo is \textit{term}},
and \enquote{tell there is a \textit{term}} is replaced by \enquote{tell there is something \textit{term}}.

The LLM response is parsed to extract all lines where text is started with a dash, taking one line as one visual feature description.

An example result of textual descriptions we obtained for the synonym set \Concept{kitty.n.04}:
\texttt{%
  \begin{itemize}
    \item Small to medium-sized mammal
    \item Furry body with distinctive patterns or colors, such as tabby, calico, or Siamese
    \item Whiskers on the face and ears
    \item  Pointed ears
    \item Large eyes with vertical pupils
    \item A short, rounded snout
    \item Retractable claws
    \item Tail that is long and fluffy, or short and stubby
    \item Often depicted in a sitting or lying position, as cats are known for their love of napping
  \end{itemize}
}

\subsubsection{Hyperparameter Study: Comparison of Text Augmenters}
\begin{figure*}
  \centering
  \includegraphics[width=.8\linewidth]{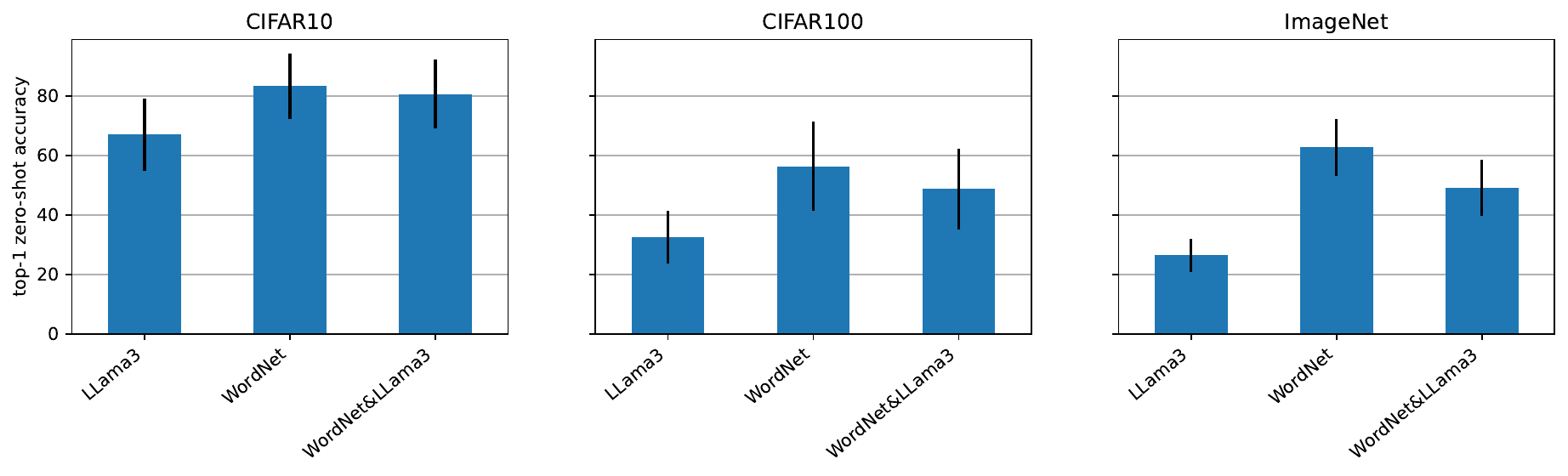}%
  \\%
  \includegraphics[width=.8\linewidth]{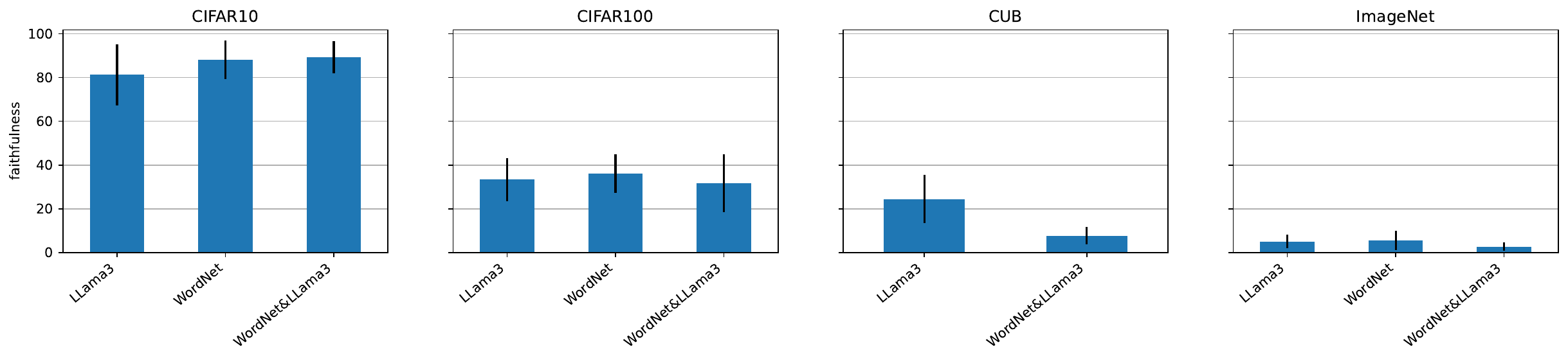}%
  \\%
  \includegraphics[width=.5\linewidth]{figures/generated/onto_m2_by_augmenter}%
  \caption{Comparison of zero-shot performance and faithfulness for textual leaf concepts and different augmenters (top); comparison of hierarchical consistency for different augmenters (bottom);
  averaged over datasets, models, (and modalities).}
  \label{fig:augmenter-choice}
\end{figure*}

We conducted a comparison of LLM-only, WordNet-only, and combined sources for textual descriptions. This attested the LLM-only version the inferior performance, indicating that descriptions of solely visual features are not sufficient for disambiguition of a concept.
Reported results therefore consider \emph{WordNet}-only text augmentations. These slightly outperformed \emph{WordNet+LLM}-combined lists of textual descriptions for each concept using Llama3 \cite{grattafiori2024llama}\footnote{\url{https://ollama.com/library/llama3}}
to generate visual feature descriptions.
For embedding a leaf concept we also compare the three modes of text-only, image-only, and average embedding thereof.
The comparison is plotted in \cref{fig:augmenter-choice}.

\subsection{Details on Ontology Manipulation \& Hyperparameter Study}
In this section, we state additional details about the UMAP based embedding transformation. Then, we present additional quantitative results for the hyperparameter search and the additional tasks for ontology manipulation, i.e. Modality alignment: steering embeddings towards the tree extracted from their leave nodes' text embeddings as well as Ontological commitment: steering them towards the closest valid SUMO tree. Furthermore, we give visual examples of resulting ontology graphs and their embedding structure based on an additional UMAP visualization (now in the usual 2D setting).

\paragraph{Parametric UMAP transformation.} To fit our self-mapping embedding transformation $H_F$ for a training set $V_\mathrm{orig} \subset \mathbb{R}^n$ and a VLM $F$, we first assign target representations initialized as the identities. Custom pairwise distances are then recomputed based on the original Euclidean distances, distances between class representation embeddings $e(\Ontology)$, and distances defined by a target class hierarchy $T'$ as assigned in Equation (5). Next, we use UMAP to rearrange the embeddings using a variation, where after each update step, all embeddings are projected onto the unit hyper-sphere. This results in a transformation $H_F^*: V_{\mathrm{orig}}\longrightarrow V_\mathrm{UMAP}$. Finally, for a transformation that generalizes to $\mathbb{R}^n$, we train a parametric regressor to approximate $H_F^*$ using a fully connected neural network with $2$ hidden layers of $n$ neurons each, ReLU activation, and $500$ training epochs.

\paragraph{Hyperparameter study and leaf swapping.} We fitted our hyperparameters for the task of steering an embedding space towards a class hierarchy $T'$ which has two swapped leave nodes. This task serves as a reference task. The hyperparameters are explored for each combination of $\alpha_\mathrm{orig} \in \{0.5, 1.0, 2.0\}$, $\beta_\mathrm{onto} \in \{ 0.5, 1.0, 2.0\}$, $\gamma_\mathrm{midp} \in \{ 0.5, 1.0, 2.0\}$. We set $N_\mathrm{UMAP} = 250$, which equals the number of samples per class if the classes are balanced in the random training subset. We keep $\mathrm{min}_\mathrm{d}=0.1$ with the default value of UMAP and the parametric model setup remains fixed. For each of the four considered VLMs ALIGN, CLIP with ViT-L14@336 backbone, SigLIP-224 and FLAVA, and each parameter configuration, we randomly swap two non-sibling nodes with positive nUTED and choose the final hyperparameter setting, which we use for all other experiments, based on a combined aggregation score $\Delta\Ontology-\text{ZS}_{\makebox[0pt][l]{\text{\scriptsize midp}}}^{\text{\scriptsize UMAP}}$. This is as stated in the main paper minimized by $\alpha_\mathrm{orig}=2,\beta_\mathrm{onto}=1, \gamma_\mathrm{midp}=2, N_\mathrm{UMAP}=250$. Numerical results for the best hyperparameter setting including all variances are presented in \cref{tab:ontology-manipulation-all-tasks}. 


\begin{table*}
  \centering
  \sisetup{table-auto-round, table-format=1.2(2), uncertainty-mode=separate}
  \small
  \begin{tabular}{@{}lrrrrrr@{}}
    \toprule
    Model & \makecell{$\Delta\Ontology$ \\ $\downarrow$} & \makecell{nUTED \\ $\downarrow$} & \makecell{ $\text{ZS}_{\makebox[0pt][l]{\text{\scriptsize text}}}^{\text{\scriptsize UMAP}}$ \\ $\uparrow$ } & \makecell{ $\text{ZS}_{\makebox[0pt][l]{\text{\scriptsize midp}}}^{\text{\scriptsize UMAP}}$ \\ $\uparrow$ } & \makecell{ $\text{ZS}_{\makebox[0pt][l]{\text{\scriptsize text}}}^{\text{\scriptsize Orig}}$ \\ $\uparrow$ } & \makecell{ $\text{ZS}_{\makebox[0pt][l]{\text{\scriptsize midp}}}^{\text{\scriptsize Orig}}$ \\ $\uparrow$ } \\
    \midrule
    \multicolumn{7}{c}{Validation task}\\
    \midrule
    ALIGN & 0.589 $\pm$ 2.9e-01 & 0.080 $\pm$ 3.7e-03 & 0.579 $\pm$ 1.3e-03 & 0.701 $\pm$ 1.5e-04 & 0.753 $\pm$ 0.0e+00 & 0.804 $\pm$ 4.1e-06 \\
    C V-L14@336 & 0.850 $\pm$ 2.8e-01 & 0.098 $\pm$ 7.1e-04 & 0.732 $\pm$ 4.5e-03 & 0.858 $\pm$ 1.5e-04 & 0.946 $\pm$ 0.0e+00 & 0.958 $\pm$ 4.0e-07 \\
    FLAVA & 0.731 $\pm$ 3.9e-01 & 0.072 $\pm$ 2.2e-03 & 0.688 $\pm$ 1.1e-03 & 0.786 $\pm$ 1.4e-04 & 0.890 $\pm$ 0.0e+00 & 0.923 $\pm$ 3.4e-07 \\
    SIGLIP 224 & 0.720 $\pm$ 8.6e-02 & 0.102 $\pm$ 3.7e-04 & 0.723 $\pm$ 2.4e-03 & 0.837 $\pm$ 6.4e-05 & 0.914 $\pm$ 0.0e+00 & 0.931 $\pm$ 1.7e-06 \\
    \midrule
    \multicolumn{7}{c}{Modality alignment}\\
    \midrule
    ALIGN & 0.723 $\pm$ 7.0e-02 & 0.189 $\pm$ 1.9e-03 & 0.547 $\pm$ 1.6e-03 & 0.674 $\pm$ 1.2e-04 & 0.753 $\pm$ 0.0e+00 & 0.803 $\pm$ 6.2e-06 \\
    C V-L14@336 & 0.779 $\pm$ 2.2e-03 & 0.105 $\pm$ 0.0e+00 & 0.782 $\pm$ 9.9e-04 & 0.871 $\pm$ 8.9e-05 & 0.946 $\pm$ 0.0e+00 & 0.958 $\pm$ 4.7e-07 \\
    FLAVA & 0.129 $\pm$ 3.1e-02 & 0.021 $\pm$ 8.3e-04 & 0.720 $\pm$ 2.2e-04 & 0.809 $\pm$ 2.7e-05 & 0.890 $\pm$ 0.0e+00 & 0.923 $\pm$ 2.1e-06 \\
    SigLIP 224 & 1.055 $\pm$ 1.3e-02 & 0.211 $\pm$ 0.0e+00 & 0.687 $\pm$ 1.9e-03 & 0.818 $\pm$ 2.8e-05 & 0.914 $\pm$ 0.0e+00 & 0.931 $\pm$ 2.1e-06 \\
    \midrule
    \multicolumn{7}{c}{Ontological commitment}\\
    \midrule
    ALIGN & 0.792 $\pm$ 2.8e-02 & 0.393 $\pm$ 5.5e-04 & 0.554 $\pm$ 7.7e-04 & 0.682 $\pm$ 1.1e-04 & 0.753 $\pm$ 0.0e+00 & 0.803 $\pm$ 5.3e-06 \\
    C V-L14@336 & 0.732 $\pm$ 1.4e-02 & 0.411 $\pm$ 1.3e-03 & 0.730 $\pm$ 1.8e-03 & 0.842 $\pm$ 2.5e-04 & 0.946 $\pm$ 0.0e+00 & 0.958 $\pm$ 4.0e-07 \\
    FLAVA & 0.714 $\pm$ 2.6e-02 & 0.400 $\pm$ 2.4e-03 & 0.671 $\pm$ 4.5e-04 & 0.781 $\pm$ 3.5e-05 & 0.890 $\pm$ 0.0e+00 & 0.923 $\pm$ 1.8e-06 \\
    SigLIP 224 & 0.889 $\pm$ 1.1e-02 & 0.411 $\pm$ 1.3e-03 & 0.745 $\pm$ 3.6e-04 & 0.820 $\pm$ 2.5e-05 & 0.914 $\pm$ 0.0e+00 & 0.931 $\pm$ 1.8e-06 \\
    \bottomrule
  \end{tabular}
  \caption{Results of the ontology manipulation: Hyperparameter search based on leaf swapping, averaged over 10 swaps and 3 runs (Validation task), adaptation to an ontology extracted from the text embeddings (Modality alignment) and to three SUMO ontologies (Ontological commitment); both averaged over 5 runs.}
  \label{tab:ontology-manipulation-all-tasks}
\end{table*}
Based on the best hyperparameters, we ran the experiment per model for 10 random leaf swaps with positive UTED and evaluate its performance on 10000 test samples. To calculate the reference mid point per class, the $2500$ randomly subsampled training samples are used. The mean results are presented and discussed in the main paper. For completeness, we list the results including variances in \cref{tab:ontology-manipulation-all-tasks}.

\begin{figure*}
  \scriptsize
  \centering
  \begin{subfigure}[t]{0.45\textwidth}
    \centering
    \begin{forest}
      [ Node 18, root node
        [ Node 17
          [ horse, leaf node ]
          [Node 15
            [ Node 10
              [ cat, leaf node ]
            [ dog, leaf node ]]
            [ Node 13
              [ frog, leaf node ]
              [ Node 11
                [ bird, leaf node ]
        [ deer, leaf node ]]]]]
        [ Node 16
          [ Node 12
            [ automobile, leaf node ]
          [ truck, leaf node ]]
          [ Node 14
            [ airplane, leaf node ]
      [ ship, leaf node ]]]]
    \end{forest}
    \caption{Ontology graph extracted from the embedded testset.}
    \label{fig:ontology_testemebddings-1}\hfill
  \end{subfigure}
  \begin{subfigure}[t]{0.45\textwidth}
    \centering
    \begin{forest}
      [ Node 18, root node
        [ Node 14
          [ dog, leaf node ]
          [ Node 12
            [ deer, leaf node ]
        [ horse, leaf node ]]]
        [ Node 17
          [ Node 15
            [ bird, leaf node ]
            [ Node 13
              [ cat, leaf node ]
          [ frog, leaf node ]]]
          [ Node 16
            [ Node 10
              [ airplane, leaf node ]
            [ ship, leaf node ]]
            [ Node 11
              [ automobile, leaf node ]
      [ truck, leaf node ]]]]]
    \end{forest}
    \caption{Ontology after learning a transformation of the embeddings.}
    \label{fig:ontology_testemebddings-2}
  \end{subfigure}

  \begin{subfigure}[t]{\textwidth}
    \centering
    \begin{forest}
      [ cow, root node
        [ canis major
          [ dog, leaf node ]
          [ equine
            [ deer, leaf node ]
        [ horse, leaf node ]]]
        [ dawdler
          [ vehicle
            [ motor vehicle
              [ automobile, leaf node ]
            [ truck, leaf node ]]
            [ jetliner
              [ airplane, leaf node ]
          [ ship, leaf node ]]]
          [ carinate
            [ bird, leaf node ]
            [ feline
              [ cat, leaf node ]
      [ frog, leaf node ]]]]]
    \end{forest}
    \caption{Target ontology graph extracted from the text embeddings.}
    \label{fig:ontology_testemebddings-3}
  \end{subfigure}
  \caption{For the flava-full VLM model we achieved to learn a transformation from the image embedding space to the text embedding space which achieves full alignment of image and text ontology (except for symmetry).}
  \label{fig:tree_steering_towards_text}
\end{figure*}
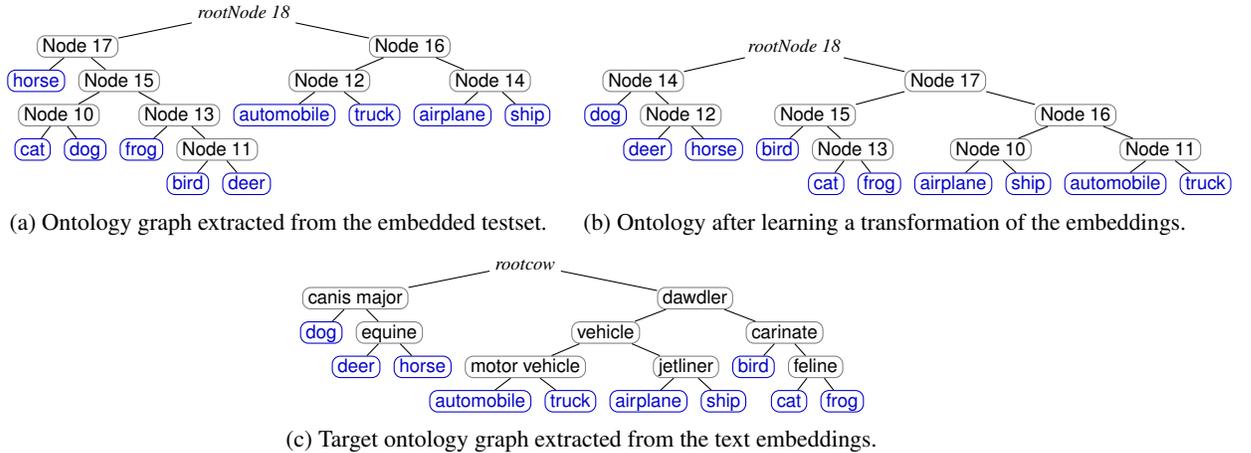

\begin{figure}[tb]
  \centering
  \includegraphics[width=0.85\textwidth]{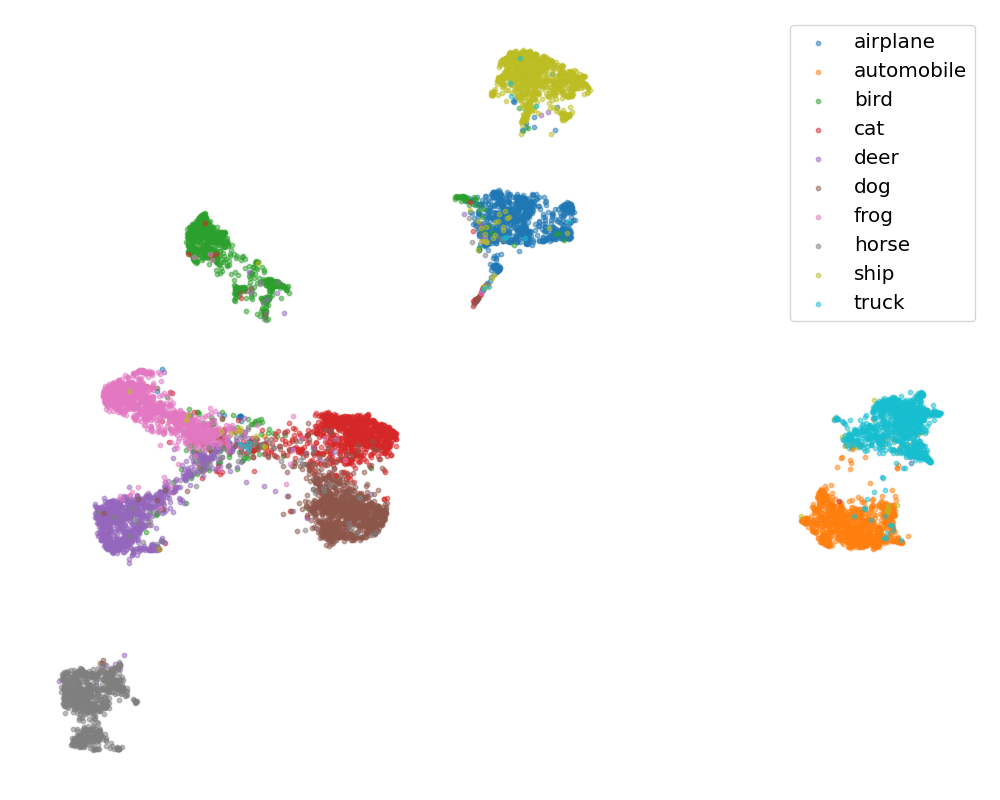}
  \caption{UMAP visualization of the original embedding space based on the test dataset}
  \label{fig:vis_emb_original_umap}
\end{figure}

\begin{figure}[tb]
  \centering
  \includegraphics[width=0.85\textwidth]{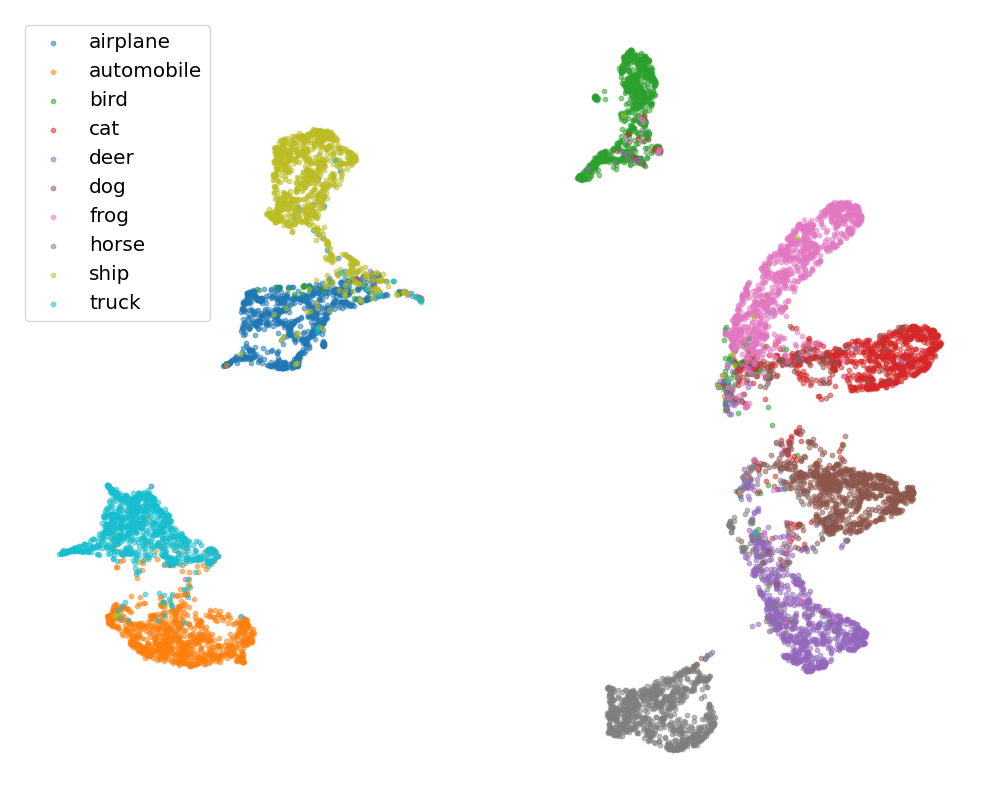}
  \caption{UMAP visualization after applying the learned transformation to the embedding space of the test dataset }
  \label{fig:vis_emb_paraumap_umap}
\end{figure}

\paragraph{Steering towards text hierarchy.} For the second task, we steer the class hierarchies of image embeddings towards hierarchies extracted from their respective text embeddings. The results, averaged over 5 transformation runs and evaluated on 10000 test samples, are presented in \cref{tab:ontology-manipulation-all-tasks}. Here, the results are mixed: While the transformations did not reduce the average graph distance to the target hierarchies for SigLIP-224, the distance is reduced substantially for FLAVA. As before, reduction in midpoint-based zero-shot performance is moderate. In addition to numerical results, we illustrate the class hierarchy manipulation for a FLAVA transformation both in form of the extracted hierarchical graph representations of original, text-target and transformed embeddings in \cref{fig:tree_steering_towards_text} and standard UMAP visualizations of the original and transformed embeddings in \cref{fig:vis_emb_original_umap,fig:vis_emb_paraumap_umap}. With the help of the transformation, the embeddings of cat and dog are now further apart and cat and frog moved closer together. Meanwhile the already aligned embeddings have not significantly changed. Except for symmetry, the extracted tree from these transformed embeddings fully aligns with the target tree extracted from the FLAVA text embeddings.

\paragraph{Steering towards closest valid hierarchies.} In our final manipulation experiment, we consider the class hierarchies from the 3 valid SUMO ontologies which are the closest to the extracted ontology from the original model embedding for all models. To steer the embeddings towards each of those, we fit a parametric UMAP transformation three times and calculate the mean per model. Numerical results are displayed in \cref{tab:ontology-manipulation-all-tasks}. On average, for all four models, we can reduce the graph distance to valid trees. However, again, we observe a tradeoff to the (moderate) reduction in midpoint-based zero-shot performance.

\paragraph{Linear Transformation}
We replace the non-linear transformation by a linear model and repeat the three tasks: validation, modality alignment and ontological commitment. In \cref{tab:lineartrafo}, we show the results. We observe (1) non-linear is in general better in achieving the target ontology with respect to nUTED and $\Delta\Ontology$ (except for SIGLIP); (2) the linear and the non-linear transformation often achieve similar image embedding zero-shot however, the linear transformation shows distictly worse text embedding zero-shot performance compared to the non-linear transformation.

\begin{table*}
  \sisetup{table-auto-round, table-format=1.2(2), uncertainty-mode=separate}
  \small
  \begin{tabular}{lllllll}
    \toprule
    Model & \makecell{$\Delta\Ontology$ \\ $\downarrow$} & \makecell{nUTED \\ $\downarrow$} & \makecell{ $\text{ZS}_{\makebox[0pt][l]{\text{\scriptsize text}}}^{\text{\scriptsize UMAP}}$ \\ $\uparrow$ } & \makecell{ $\text{ZS}_{\makebox[0pt][l]{\text{\scriptsize midp}}}^{\text{\scriptsize UMAP}}$ \\ $\uparrow$ } & \makecell{ $\text{ZS}_{\makebox[0pt][l]{\text{\scriptsize text}}}^{\text{\scriptsize Orig}}$ \\ $\uparrow$ } & \makecell{ $\text{ZS}_{\makebox[0pt][l]{\text{\scriptsize midp}}}^{\text{\scriptsize Orig}}$ \\ $\uparrow$ } \\
    \midrule
    \multicolumn{7}{c}{Validation task}\\
    \midrule
    ALIGN & 0.812 $\pm$ 1.7e-01 & 0.108 $\pm$ 2.0e-03 & 0.485 $\pm$ 1.6e-03 & 0.715 $\pm$ 1.3e-04 & 0.753 $\pm$ 0.0e+00 & 0.803 $\pm$ 5.1e-06 \\
    C V-L14@336 & 0.927 $\pm$ 2.6e-01 & 0.103 $\pm$ 2.2e-04 & 0.595 $\pm$ 4.3e-03 & 0.887 $\pm$ 9.9e-05 & 0.946 $\pm$ 0.0e+00 & 0.958 $\pm$ 3.8e-07 \\
    FLAVA & 0.799 $\pm$ 3.6e-01 & 0.090 $\pm$ 2.0e-03 & 0.712 $\pm$ 8.8e-04 & 0.850 $\pm$ 9.2e-05 & 0.890 $\pm$ 0.0e+00 & 0.923 $\pm$ 1.7e-06 \\
    SigLIP 224 & 0.547 $\pm$ 1.9e-01 & 0.082 $\pm$ 1.9e-03 & 0.623 $\pm$ 4.5e-03 & 0.861 $\pm$ 6.3e-05 & 0.914 $\pm$ 0.0e+00 & 0.931 $\pm$ 1.7e-06 \\
    \midrule
    \multicolumn{7}{c}{Modality alignment}\\
    \midrule
    ALIGN & 1.013 $\pm$ 0.0e+00 & 0.237 $\pm$ 0.0e+00 & 0.510 $\pm$ 3.9e-04 & 0.693 $\pm$ 1.2e-04 & 0.753 $\pm$ 0.0e+00 & 0.803 $\pm$ 6.2e-06 \\
    C V-L14@336 & 0.748 $\pm$ 1.1e-03 & 0.105 $\pm$ 0.0e+00 & 0.690 $\pm$ 2.0e-03 & 0.895 $\pm$ 4.5e-05 & 0.946 $\pm$ 0.0e+00 & 0.958 $\pm$ 4.7e-07 \\
    FLAVA & 0.443 $\pm$ 4.4e-03 & 0.295 $\pm$ 8.3e-04 & 0.729 $\pm$ 5.1e-04 & 0.868 $\pm$ 1.7e-05 & 0.890 $\pm$ 0.0e+00 & 0.923 $\pm$ 2.1e-06 \\
    SigLIP 224 & 1.055 $\pm$ 1.3e-02 & 0.211 $\pm$ 0.0e+00 & 0.625 $\pm$ 3.2e-03 & 0.846 $\pm$ 4.6e-05 & 0.914 $\pm$ 0.0e+00 & 0.931 $\pm$ 2.1e-06 \\
    \midrule
    \multicolumn{7}{c}{Ontological commitment}\\
    \midrule
    ALIGN & 0.890 $\pm$ 1.7e-02 & 0.418 $\pm$ 2.5e-03 & 0.497 $\pm$ 6.3e-04 & 0.700 $\pm$ 7.8e-05 & 0.753 $\pm$ 0.0e+00 & 0.803 $\pm$ 5.3e-06 \\
    C V-L14@336 & 0.778 $\pm$ 1.4e-02 & 0.422 $\pm$ 3.3e-03 & 0.601 $\pm$ 4.7e-03 & 0.872 $\pm$ 2.0e-04 & 0.946 $\pm$ 0.0e+00 & 0.958 $\pm$ 4.0e-07 \\
    FLAVA & 0.813 $\pm$ 3.7e-02 & 0.402 $\pm$ 2.8e-03 & 0.709 $\pm$ 2.2e-04 & 0.848 $\pm$ 1.7e-05 & 0.890 $\pm$ 0.0e+00 & 0.923 $\pm$ 1.8e-06 \\
    SigLIP 224 & 0.889 $\pm$ 1.1e-02 & 0.411 $\pm$ 1.3e-03 & 0.641 $\pm$ 2.5e-03 & 0.849 $\pm$ 3.3e-05 & 0.914 $\pm$ 0.0e+00 & 0.931 $\pm$ 1.8e-06 \\
    \bottomrule
  \end{tabular}
  \caption{{Simple Linear Model to approximate the denoted embedding transformation, all runs are averaged over five seeds.}\label{tab:lineartrafo}
  }
\end{table*}




